\documentclass{article}

\usepackage[numbers]{natbib}

\usepackage[final]{neurips_2024}

\usepackage[utf8]{inputenc} % allow utf-8 input
\usepackage[T1]{fontenc}    % use 8-bit T1 fonts
\usepackage{hyperref}       % hyperlinks
\usepackage{url}            % simple URL typesetting
\usepackage{booktabs}       % professional-quality tables
\usepackage{amsfonts}       % blackboard math symbols
\usepackage{nicefrac}       % compact symbols for 1/2, etc.
\usepackage{microtype}      % microtypography
\usepackage{xcolor}         % colors

\usepackage{subfigure}
\usepackage{subfigure,graphicx,dsfont,amssymb,amsmath,calc,tabulary}

\usepackage{amsmath}
\usepackage{amssymb}
\usepackage{amsthm}
\usepackage{fixmath}
\usepackage{algorithm}
\usepackage{algorithmic}
\newtheorem{theorem}{Theorem}

\usepackage{wrapfig}

\usepackage{amsmath,amsfonts,bm}

\def\Algref#1{Algorithm~\ref{#1}}

\def\1{\bm{1}}

\def\vtheta{{\bm{\theta}}}
\def\va{{\bm{a}}}

\def\vx{{\bm{x}}}

\def\vz{{\bm{z}}}

\DeclareMathAlphabet{\mathsfit}{\encodingdefault}{\sfdefault}{m}{sl}
\SetMathAlphabet{\mathsfit}{bold}{\encodingdefault}{\sfdefault}{bx}{n}

\def\gL{{\mathcal{L}}}

\def\gO{{\mathcal{O}}}

\def\sA{{\mathbb{A}}}
\def\sB{{\mathbb{B}}}

\def\sD{{\mathbb{D}}}
\def\sH{{\mathbb{H}}}

\def\sM{{\mathbb{M}}}

\def\sR{{\mathbb{R}}}

\newcommand{\E}{\mathbb{E}}

\DeclareMathOperator*{\argmin}{arg\,min}

\title{Personalized Federated Learning with\\ Mixture of Models for Adaptive Prediction and\\ Model Fine-Tuning}

\author{%
  Pouya M.~Ghari \\
  University of California Irvine\\
  \texttt{pmollaeb@uci.edu}
  \And
  Yanning Shen \thanks{Corresponding author} \\
  University of California Irvine\\
  \texttt{yannings@uci.edu}
}

\begin{document}

\maketitle

\begin{abstract}
  % In an online federated learning framework, a set of clients collaborates with a server to train a model online. Specifically, at each time step $t$, clients receive a global model from the server with which they make predictions for a data stream they receive. Clients incur losses after making the predictions and they update the global model locally based on the incurred losses. Aggregating updates from clients, the server updates the global model. Then the updated global model is transmitted to and employed by clients in the next time step. This continues up until the time horizon $T$. Employing existing online federated learning algorithms guarantee regret upper bound of $\gO(N\sqrt{T})$ where $N$ is the number of clients. The present paper proves that it is possible to obtain improved regret bound of $\gO(\sqrt{NT})$ for online  federated learning. In particular, the present paper proposes a novel online federated learning algorithm for cases where the model can be expressed as a linear function of the input data representation. It is proved that the proposed algorithm achieves regret of $\gO(\sqrt{NT})$. Experiments on real datasets demonstrate the effectiveness of the proposed algorithm.
  Federated learning is renowned for its efficacy in distributed model training, ensuring that users, called clients, retain data privacy by not disclosing their data to the central server that orchestrates collaborations. Most previous work on federated learning assumes that clients possess static batches of training data. However, clients may also need to make real-time predictions on streaming data in non-stationary environments. In such dynamic environments, employing pre-trained models may be inefficient, as they struggle to adapt to the constantly evolving data streams. To address this challenge, clients can fine-tune models online, leveraging their observed data to enhance performance. Despite the potential benefits of client participation in federated online model fine-tuning, existing analyses have not conclusively demonstrated its superiority over local model fine-tuning. To bridge this gap, the present paper develops a novel personalized federated learning algorithm, wherein each client constructs a personalized model by combining a locally fine-tuned model with multiple federated models learned by the server over time. Theoretical analysis and experiments on real datasets corroborate the effectiveness of this approach for real-time predictions and federated model fine-tuning.  
\end{abstract}

\section{Introduction}
Federated learning enables a group of learners, known as \emph{clients}, to collaborate and collectively train a model under the coordination of a central \emph{server}, without revealing their data. In this framework, clients perform local model updates and share these updates with the server. By aggregating these local updates, the server globally updates the model. Many prior works in the literature have assumed that each client stores a batch of training data and updates models locally based on this stored data (see e.g., \cite{Mohri2019,Li2020,Rothchild2020,Charles2021,Marfoq2021,Wang2022}). However, in some cases, clients may need to make real-time predictions, and streams of data arrive sequentially, making it challenging to store and process data in batch. Furthermore, if clients operate in a non-stationary and dynamic environment, employing pre-trained models may fall short in prediction accuracy, requiring clients to fine-tune their models to adapt to their data. 

A federated learning framework, commonly referred to as online federated learning \cite{Mitra2021, Hong2022, Ghari2022}, specifically addresses situations where clients engage in real-time predictions using a shared global model. After making predictions, clients collaborate with the server to update the global model for subsequent predictions in the future. Specifically, after making predictions, clients incur losses and based on the incurred losses, clients update the model locally and send their local updates to the server. The server updates the global model upon aggregating local updates and then distributes the updated global model to clients for use in the ongoing online prediction task. In this context, the performance of clients can be assessed using the notion of \emph{regret} \cite{Cesa-Bianchi2006,Auer2003}. The regret of a client at a given time step is defined as the difference between its prediction loss and that of the best model in hindsight. The best model in hindsight is determined as the model that achieves the minimum total prediction loss over time across all clients' data. The primary objective is to minimize the cumulative regret of all clients over time.

In the literature, federated learning algorithms based on online gradient descent have been proposed that achieve sub-linear regret upper bounds \cite{Mitra2021,Hong2022,Ghari2022}. This suggests that over the long run, these algorithms perform as well as the best model in hindsight. However, in Section \ref{OFGD}, this paper demonstrates that these federated learning algorithms do not obtain tighter regret bounds compared to the scenario where each client learns its own model locally without participating in federated learning. This indicates that participation in federated learning may not provide any benefit for clients performing online prediction. Specifically, if data is distributed non-i.i.d. among clients and data distributions are time-variant and not known a priori, it is likely that local model training by clients will achieve better prediction accuracy than participation in federated learning. Although the benefit of federated learning in online prediction is not evident in existing theoretical analyses, models learned through federated learning enjoy higher generalizability as they are trained on all data samples distributed among clients. This motivates the idea that combining the model learned through federated learning with the locally learned one may prove effective in scenarios where clients need to perform online prediction. %However, the regret upper bounds of these algorithms linearly scales with the number of clients. Thus, if each client opts not to collaborate in federated learning and instead learns its model locally and independently using online convex optimization techniques, the cumulative regret of clients would be the same as the regret upper bounds found in \cite{Mitra2021,Hong2022,Ghari2022}. Therefore, the benefit of collaboration among clients is not reflected in regret upper bounds of existing online federated learning algorithms. 

This paper proposes the Fed-POE (Federated Learning with Personalized Online Ensemble) algorithm, through which each client constructs a personalized model for online prediction by adaptively ensembling the locally learned model and the model learned through federated learning. With Fed-POE, clients can adapt their local models to their data while benefiting from the higher generalizability of the federated model. Theoretical analysis for convex cases demonstrates that Fed-POE achieves sublinear regret bounds with respect to both the best federated and local models in hindsight. This indicates that Fed-POE effectively leverages the advantages of both federated and local models. Providing such theoretical guarantees may not be feasible for non-convex models such as neural networks. These models may suffer from the forgetting process \cite{Toneva2019, Ramasesh2022}, where fine-tuning on streaming data `on the fly' can lead to forgetting previously observed data samples. To overcome this challenge, the present paper proposes a novel framework in which the server periodically stores federated model parameters over time. Each client adaptively selects a personalized subset of stored models on the server based on the performance of these models in the client's online prediction task. Clients then use the selected models, along with the federated and local models, to construct an ensemble model for prediction. Clients select a subset of models to both control the memory and computational complexity of prediction and to prune models with relatively lower accuracy, thereby improving prediction performance. Theoretical analysis proves that Fed-POE achieves sublinear regret with respect to the best model in hindsight among the local model, federated model, and all models stored by the server. The contributions of the present paper are summarized as follows:
\begin{itemize}
    \item Fed-POE enables clients to utilize the advantages of both local and federated models for online prediction tasks.
    \item To address the issue of forgetting in online prediction, Fed-POE introduces a novel federated framework for collaboration between clients and the server.
    \item Theoretical analyses for both convex and non-convex cases prove that Fed-POE achieves sublinear regret with respect to the best model in hindsight.
    \item Extensive experiments on regression and image classification datasets show that Fed-POE effectively leverages the benefits of local and federated models, achieving higher online prediction accuracy compared to state-of-the-art federated learning algorithms.
\end{itemize}

\section{Related Works} \label{others}
\textbf{Personalized Federated Learning.} Personalized federated learning involves developing individualized models for each client, derived from a global model learned through federated learning. Several personalized federated learning algorithms have been proposed in the literature \cite{Tan2022, Chen2022, Wu2023, Ye2023}.  In \cite{Hanzely2020, Dinh2020, TLi2021, Liu2022}, clients construct their personalized models by adding a regularization term to the local objective and using the global federated model. Algorithms in \cite{Fallah2020, Acar2021} allow clients to fine-tune the global federated model using model-agnostic meta-learning \cite{Finn2017} to learn their personalized models. With Fed-Rep \cite{Collins2021}, each client generates a representation using the global model and learns its own local head for prediction. Algorithms in \cite{Deng2020, LiX2021, Zhang2023} enable clients to achieve personalized models by combining local and global models. However, none of these works have addressed the problem of online prediction while clients collaborate on training their personalized models.

\textbf{Federated Learning with Streaming Data.} Several studies have explored the problem of federated learning when clients receive a stream of data in real time. While \cite{Chen2020, Mawuli2023, Marfoq2023} investigate federated learning scenarios where clients receive new training data in each learning round, they do not address the aspect of online prediction by clients. Consequently, these works cannot provide regret guarantees for online prediction. Additionally, \cite{Damaskinos2022} studies the issue of staleness in online federated learning to both train and make prediction with the model; however, it lacks rigorous theoretical analysis. In \cite{Mitra2021}, an online federated learning algorithm is introduced, utilizing online mirror descent, with a proven sublinear regret bound. Similarly, \cite{Park2022,Ganguly2022} propose online federated learning algorithms with guaranteed sublinear regret. Furthermore, \cite{Patel2023} analyzes the benefits of collaboration in online federated learning, particularly in scenarios where only loss values at queried points are available to clients, without access to loss gradients. Regarding online federated kernel learning, algorithms are introduced in \cite{Gauthier2022,Gogineni2023}, albeit without accompanying regret analysis, leading to an absence of guaranteed regret bounds. In \cite{Hong2022,Ghari2022}, multiple kernel-based models and random feature-based online federated kernel learning algorithms are proposed, with demonstrated sublinear regret.

\section{Preliminaries} \label{OFGD}
This section explains the problem of federated learning for real-time prediction and model training. The present section studies the cases where clients either collaborate in federated learning or employ online gradient descent methods to locally train the model in real-time. This study highlights the motivation behind the proposed ensemble approach to federated learning.%, particularly emphasizing its relevance in scenarios where data samples are distributed heterogeneously among clients.

\subsection{Online Prediction and Federated Learning}
Let there are $N$ clients interact with a server to train a model $f(\cdot;\cdot)$. Also, let $[N]:=\{1,\ldots,N\}$. At each time step $t$, client $i$, $\forall i \in [N]$ receives a data sample $\vx_{i,t} \in \sR^d$ and makes the prediction $f(\vx_{i,t};\vtheta_t)$ where $\vtheta_t$ denotes the parameter of the model at time step $t$. Note that generalization to the scenario where at each time step, each client receives a dataset instead of a single data sample is straightforward. After making prediction, client $i$, $\forall i \in [N]$ observes label $y_{i,t}$. Then client $i$, $\forall i \in [N]$ computes the loss of its prediction $\gL(f(\vx_{i,t};\vtheta_t),y_{i,t})$ where $\gL(\cdot,\cdot)$ denotes the loss function. After computing the loss, clients send their updates to the server (e.g., by sending the gradient loss $\nabla \gL(f(\vx_{i,t};\vtheta_t),y_{i,t})$) and the server aggregates information from clients to update $\vtheta_t$ to $\vtheta_{t+1}$ to be used by clients to make predictions at time step $t+1$. For ease of presentation, it is assumed that all clients can send their updates to the server every time step. Generalizing the results for the cases where only a fraction of clients can send their updates is straightforward. Furthermore, it is assumed that the label $y_{i,t}$ for any $i$ and $t$ is determined by the environment through a process unknown to the clients. This implies that the data distribution observed by client $i$ can be non-stationary, and client $i$, $\forall i \in [N]$ does not know the distribution. Additionally, the data distribution can differ across clients. The goal is to enable clients to collaborate with the server to minimize the cumulative regret of clients over time. The regret of client $i$ at time step $t$ is defined as the difference between the prediction loss $\gL(f(\vx_{i,t};\vtheta_t),y_{i,t})$ and the loss $\gL(f(\vx_{i,t};\vtheta^*),y_{i,t})$ corresponding to the model with the optimal parameter $\vtheta^*$. Therefore, the average cumulative regret of clients up to time horizon $T$ is defined as
\begin{align}
    R_T = \frac{1}{N}\sum_{t=1}^{T}{\sum_{i=1}^{N}{\gL(f(\vx_{i,t};\vtheta_t),y_{i,t})}} - \frac{1}{N}\sum_{t=1}^{T}{\sum_{i=1}^{N}{\gL(f(\vx_{i,t};\vtheta^*),y_{i,t})}} \label{eq:1}
\end{align}
where $\vtheta^*$ denotes the optimal model parameter in hindsight and can be expressed as
\begin{align}
    \vtheta^* = \argmin_{\vtheta} \sum_{t=1}^{T}{\sum_{i=1}^{N}{\gL(f(\vx_{i,t};\vtheta),y_{i,t})}}. \label{eq:2}
\end{align}
One common way to solve the problem is that clients employ online gradient descent to update models locally and the server exploits federated averaging \cite{Li2020b,Mitra2021} to update the global model.

\subsection{Federated Learning with Online Gradient Descent} \label{ogd}
%At each time step $t$, the server constructs the set of $M$ clients $\sC_t$ by choosing $M$ clients uniformly at random if $M<N$. Let $\rho$ denote the probability that a client is chosen to be in $\sC_t$. Therefore, it can be inferred that $\rho = \min(\frac{M}{N},1)$. 
At each time step $t$, client $i$ obtains the locally updated model parameter $\boldsymbol{\psi}_{i,t+1}$ as follows:
\begin{align}
    \boldsymbol{\psi}_{i,t+1} = \vtheta_t - \eta \nabla \gL(f(\vx_{i,t};\vtheta_t),y_{i,t}) \label{eq:3}
\end{align}
where $\eta$ is the learning rate. Aggregating locally updated model parameters, the server obtains the updated global model parameter for time step $t+1$ as
\begin{align}
    \vtheta_{t+1} = \frac{1}{N}\sum_{i=1}^{N}{\boldsymbol{\psi}_{i,t+1}}. \label{eq:4}
\end{align}
This continues up until the time horizon $T$. This paper examines regret under some or all of the following assumptions:\\
% \begin{itemize}
    % \item[\textbf{A1.}] 
    \textbf{A1.} The loss function $\gL(f(\vx;\vtheta),y)$ is convex with respect to $\vtheta$. \\
    % \item[\textbf{A2.}] The gradient of the loss is bounded as $\|\nabla \gL(f(\vx;\vtheta),y)\| \le G$.
    \textbf{A2.} The gradient of the loss is bounded as $\|\nabla \gL(f(\vx;\vtheta),y)\| \le G$. \\
    % \item[\textbf{A3.}] For any $\vx$, $\vtheta$, the loss is bounded as $0 \le \gL(f(\vx;\vtheta),y) \le 1$.
    \textbf{A3.} For any $\vx$, $\vtheta$, the loss is bounded as $0 \le \gL(f(\vx;\vtheta),y) \le 1$.
    % \item[\textbf{A3.}] The loss function $\gL(\cdot,\cdot)$ is $L_c$-Lipschitz such that
    % \begin{align}
    %     |\gL(f_1,y) - \gL(f_2,y)| \le L_c|f_1 - f_2|, \forall f_1, f_2. \nonumber
    % \end{align}
    % Furthermore, the gradient of loss function $\nabla \gL(\cdot,\cdot)$ is $L_g$-Lipschitz such that
    % \begin{align}
    %     \|\nabla\gL(f_1,y) - \nabla\gL(f_2,y)\| \le L_g|f_1 - f_2|, \forall f_1, f_2. \nonumber
    % \end{align}
% \end{itemize}

%In this paper, $\|\cdot\|_p$ denotes $\ell_p$-norm while for the sake of notation $\|\cdot\|$ denotes $\ell_2$-norm. 
The following theorem specifies the regret bound for federated learning employing online gradient descent under A1 and A2. 

\begin{theorem} \label{th:1}
Employing online gradient descent, the following cumulative regret upper bound is guaranteed for federated learning under assumptions A1 and A2:
\begin{align}
    \frac{1}{N}\sum_{t=1}^{T}{\sum_{i=1}^{N}{\gL(f(\vx_{i,t};\vtheta_t),y_{i,t})}} - \frac{1}{N}\sum_{t=1}^{T}{\sum_{i=1}^{N}{\gL(f(\vx_{i,t};\vtheta^*),y_{i,t})}} \le \frac{\|\vtheta^*\|^2 }{2\eta} + \frac{\eta }{2}G^2T. \label{eq:5}
\end{align}
%where the expectation is taken with respect to randomization in choosing clients by the server.
\end{theorem}

Proof of Theorem \ref{th:1} can be found in Appendix \ref{A}. According to Theorem \ref{th:1}, choosing $\eta = \gO(\sqrt{1/T})$, the cumulative regret in \eqref{eq:5} is bounded from above as $\gO(\sqrt{T})$. If the time horizon $T$ is unknown, the doubling trick technique (see e.g., \cite{Alon2017,Ghari2024tnnls}) can be effectively used to set the learning rate to maintain theoretical guarantees. Consider the case where each client learns the model locally using online gradient descent without collaborating with other clients and the server.
Let $\boldsymbol{\phi}_{i,t}$ denote the local model parameter learned by client $i$ at time step $t$, employing online gradient descent locally. The regret of client $i$ in this case is equivalent to federated learning regret where there is only one client. Therefore, substituting $N=1$ and $\vtheta_0 = \boldsymbol{0}$ in \eqref{eq:6ap} of Appendix \ref{A}, for the cumulative regret of client $i$ with respect to any $\vtheta$, we get
\begin{align}
    \sum_{t=1}^{T}{\gL(f(\vx_{i,t};\boldsymbol{\phi}_{i,t}),y_{i,t})} - \sum_{t=1}^{T}{\gL(f(\vx_{i,t};\vtheta),y_{i,t})} \le \frac{\|\vtheta\|^2 }{2\eta} + \frac{\eta}{2}G^2T. \label{eq:71ap}
\end{align}
Averaging \eqref{eq:71ap} over all clients and substituting $\vtheta$ with $\vtheta^*$ in \eqref{eq:71ap}, we obtain
\begin{align}
    \frac{1}{N}\sum_{t=1}^{T}{\sum_{i=1}^{N}{\gL(f(\vx_{i,t};\boldsymbol{\phi}_{i,t}),y_{i,t})}} - \frac{1}{N}\sum_{t=1}^{T}{\!\sum_{i=1}^{N}{\gL(f(\vx_{i,t};\vtheta^*),y_{i,t})}} \le \frac{\|\vtheta^*\|^2 }{2\eta} + \frac{\eta }{2}G^2T. \label{eq:1r2}
\end{align}
%where the expectation is taken with respect to randomization in choosing clients by the server.
Comparing \eqref{eq:5} with \eqref{eq:1r2}, it can be inferred that under assumptions A1 and A2, federated learning does not achieve tighter regret bound than the case where each client independently learns its local model. Hence, from a theoretical standpoint, it remains uncertain whether collaboration in federated learning yields any improvement over local online model training. Intuitively, collaboration in federated learning may prove advantageous when there exists similarity among data samples observed by clients over time. However, in cases where data distribution is heterogeneous and such similarities are lacking, employing online local training may yield superior results for a client. Given the lack of prior information on data distribution in online scenarios, each client can independently assess over time whether utilizing the model learned through federated learning for predictions is beneficial.

Furthermore, according to assumption A1, the theoretical guarantees obtained in \eqref{eq:5} and \eqref{eq:1r2} hold if the loss is convex with respect to the model parameter. However, in the case of non-convex models, such as neural networks, these theoretical guarantees are not applicable. Non-convex models, like neural networks, may encounter the forgetting process \cite{Toneva2019,Ramasesh2022}, where the model tends to overfit to recently observed data samples. Consequently, it may not be feasible to derive a single model parameter using online gradient descent that achieves sublinear regret with respect to the best model in hindsight. The subsequent section introduces a novel algorithm aimed at assisting clients in addressing such scenarios.

\section{Personalized Federated Learning Methods } \label{OFRL}
The current section introduces personalized federated learning algorithms where each client dynamically learns the prediction performance of both the models trained through federated learning over time and the locally learned model.

\subsection{Ensemble Learning} \label{ens}
At each time step, each client constructs an ensemble model comprising the federated model and its locally learned model to make a prediction. Let $\boldsymbol{\phi}_{i,t}$ be the model parameter locally learned by client $i$ such that at each time step $t$, client $i$ updates $\boldsymbol{\phi}_{i,t}$ via gradient descent as
\begin{align}
    \boldsymbol{\phi}_{i,t+1} = \boldsymbol{\phi}_{i,t} - \eta \nabla \gL(f(\vx_{i,t};\boldsymbol{\phi}_{i,t}),y_{i,t}). \label{eq:6}
\end{align}
Furthermore, let clients and the server collaborate to update the \emph{federated} model parameter $\vtheta_t$ as outlined in \eqref{eq:3} and \eqref{eq:4}. At time step $t$, client $i$ makes the prediction for $\vx_{i,t}$ using its personalized ensemble model $f_{i,t}(\cdot)$ expressed as
\begin{align}
    f_{i,t}(\vx_{i,t}) = \frac{\alpha_{i,t}}{\alpha_{i,t}+\beta_{i,t}}f(\vx_{i,t};\vtheta_t) + \frac{\beta_{i,t}}{\alpha_{i,t}+\beta_{i,t}}f(\vx_{i,t};\boldsymbol{\phi}_{i,t}) \label{eq:7}
\end{align}
where $\alpha_{i,t}$ and $\beta_{i,t}$ represent the weights assigned by client $i$ to the federated and local models, respectively, indicating the credibility of predictions from each model. After making predictions and observing the label $y_{i,t}$, client $i$ computes the loss of predictions from the federated and local models, updating the weights $\alpha_{i,t}$ and $\beta_{i,t}$ using the multiplicative update rule as follows:
\begin{subequations} \label{eq:8}
\begin{align}
    \alpha_{i,t+1} &= \alpha_{i,t}\exp\left(-\eta_c \gL(f(\vx_{i,t};\vtheta_t),y_{i,t})\right), \label{eq:8a} \\
    \beta_{i,t+1} &= \beta_{i,t}\exp\left(-\eta_c \gL(f(\vx_{i,t};\boldsymbol{\phi}_{i,t}),y_{i,t})\right) \label{eq:8b}
\end{align}
\end{subequations}
where $\eta_c$ is a learning rate. Client $i$ initializes $\alpha_{i,1}=1$ and $\beta_{i,1}=1$.
The proposed algorithm is personalized since according to \eqref{eq:7}, each client constructs its own ensemble model to perform prediction. The personalized regret of client $i$ is defined as
% \begin{align}
    $C_{i,T} = \sum_{t=1}^{T}{\gL(f_{i,t}(\vx_{i,t}),y_{i,t})} - \sum_{t=1}^{T}{\gL(f(\vx_{i,t};\boldsymbol{\phi}_i^*),y_{i,t})}$ %\label{eq:10}
% \end{align} 
where $\boldsymbol{\phi}_i^*$ denotes the best hindsight model parameter for client $i$ which can be expressed as
%\begin{align}
    $\boldsymbol{\phi}_i^* = \argmin_{\boldsymbol{\phi}}{\sum_{t=1}^{T}{\gL(f(\vx_{i,t};\boldsymbol{\phi}_i),y_{i,t})}}$. %\label{eq:11}
%\end{align}
The following theorem establishes the personalized regret upper bound for client $i$ as well as global regret of all clients with respect to the best model parameter in hindsight.
\begin{theorem} \label{th:2}
     Under assumptions A1--A3, employing the ensemble model in \eqref{eq:7} for online prediction, the global regret of all clients is bounded from above as 
     \begin{align}
         \frac{1}{N}\!\sum_{t=1}^{T}\!{\sum_{i=1}^{N}{\!\gL(f_{i,t}(\vx_{i,t}),y_{i,t})}} \!-\! \frac{1}{N}\!\sum_{t=1}^{T}{\!\sum_{i=1}^{N}{\!\gL(f(\vx_{i,t};\vtheta^*),y_{i,t})}} \!\le\! \frac{\|\vtheta^*\|^2 }{2\eta} \!+\! \frac{\ln(2)}{\eta_c} \!+\! \frac{\eta }{2}G^2T \!+\! \frac{\eta_c T}{2} \label{eq:12}
     \end{align}
     while client $i$ achieves the following personalized regret upper bound:
     \begin{align}
         \sum_{t=1}^{T}{\gL(f_{i,t}(\vx_{i,t}),y_{i,t})} - \sum_{t=1}^{T}{\gL(f(\vx_{i,t};\boldsymbol{\phi}_i^*),y_{i,t})} \le \frac{\|\boldsymbol{\phi}_i^*\|^2 }{2\eta} + \frac{\ln(2)}{\eta_c} + \frac{\eta}{2}G^2T + \frac{\eta_c T}{2}. \label{eq:9}
     \end{align}
\end{theorem}

Proof of Theorem \ref{th:2} is given in Appendix \ref{B}. According to \eqref{eq:12} and \eqref{eq:9} in Theorem \ref{th:2}, setting $\eta = \gO(1/\sqrt{T})$, $\eta_c = \gO(1/\sqrt{T})$ both the global regret for all clients and the personalized regret of each client $i$ achieve a regret bound of $\gO(\sqrt{T})$. This demonstrates that while the ensemble model in \eqref{eq:7} ensures personalized regret guarantees for each client with respect to its best model in hindsight, it also enables clients to leverage federated learning, thus enjoying sublinear regret in comparison to the best global model in hindsight.

\textbf{Comparison with Federated and Local Models.} The main advantage of using the ensemble model instead of the federated model lies in Theorem \ref{th:2}, where it is shown that the ensemble model can attain the global regret guarantee provided by the federated model while employing the federated model, achieving the personalized regret guarantee in \eqref{eq:9} is not feasible. However, according to \eqref{eq:71ap} and \eqref{eq:1r2}, using the online local training, each client achieves sublinear regret with respect to its best model while all clients achieve sublinear regret with respect to the best global model in hindsight. This efficacy of local training stems from clients adapting the model to their individual data. In contrast, in federated learning, the model is trained on all data samples across clients, potentially leading to higher generalizability compared to its local counterpart. If there are similarities in the distribution of data samples among clients over time, the federated model is anticipated to achieve greater accuracy in online prediction. However, due to the lack of available information regarding the relationships between data samples observed by clients before online prediction, these advantages may not be reflected in theoretical bounds. The proposed method constructs an ensemble to harness the advantages of both federated and local models for online prediction, as evidenced by Theorem \ref{th:2}. Experimental results in Subsection in \ref{exp:reg} confirm that the ensemble model achieves superior performance compared to both local and federated models.

\subsection{Model Selection}
Regret guarantees for the ensemble method, as outlined in Theorem \ref{th:2}, are contingent upon the model's convexity. However, if the model is non-convex, achieving such guarantees may not be feasible. Particularly, non-convex models such as neural networks are susceptible to the forgetting process \cite{Toneva2019,Ramasesh2022}, wherein applying online gradient descent may lead to overfitting to recently observed data samples. This section introduces a novel algorithm that allows clients to make online predictions using non-convex models while simultaneously collaborating to fine-tune the model. The scenario assumes the existence of a pre-trained model, and the objective for clients is two-fold: to make real-time predictions and to refine the model for alignment with their preferences. This situation may arise, for instance, in fine-tuning large foundation models to tailor them to client preferences.

Let the server and clients collaborate to fine-tune the non-convex model $f(\cdot;\cdot)$. At each time step $t$, client $i$ updates the model on the batch of recently observed samples with size $b$ as
\begin{align}
    \boldsymbol{\psi}_{i,t+1} = \vtheta_t - \frac{\eta}{b} \sum_{\tau=t-b}^{t}{\nabla \gL(f(\vx_{i,\tau};\vtheta_t),y_{i,\tau})}. \label{eq:13}
\end{align}
Then the server aggregates locally updated parameters and updates the federated model parameter as in \eqref{eq:4}. Furthermore, each clients learns its own local model by fine-tuning the pre-trained model locally via online gradient descent as in \eqref{eq:6} on the batch of $b$ recently observed samples. While online gradient descent methods are well-known for their efficiency in handling dynamic environments, employing the update rule of \eqref{eq:13} for non-convex models may lead to overfitting to recently observed batches. To mitigate potential forgetting, the server saves the federated model parameters every $n$ time step, where $n$ is an integer hyperparameter. 

\begin{algorithm}[tb]
	\caption{Model selection by client $i$ at time step $t$}
	\label{alg:2}
	\begin{algorithmic}[1]
		\STATE {\bfseries Input:}{ $\sD_t$, $M$, wights $\{w_{ij,t}\}_{j=1}^{|\sD_t|}$ and $\sM_{i,t}=\varnothing$.}
		\FOR{$m=1,\ldots,M$}
            \STATE Sample model index $k$ according to the PMF $p_{ij,t} = \frac{w_{ij,t}}{\sum_{j=1}^{|\sD_t|} w_{ij,t}}$, $\forall j \in [|\sD_t|]$. \label{step:31}
            \IF{$k \notin \sM_{i,t}$}
            \STATE Append model index $k$ to $\sM_{i,t}$.
            \ENDIF
		\ENDFOR
            \STATE {\bfseries Output:}{ $\sM_{i,t}$.}
	\end{algorithmic}
\end{algorithm}
Let $\sD_t$ represent the set of model parameters stored by the server at time step $t$. At time step $\tau = (j-1)n + 1$, the server adds $\vtheta_\tau$ to $\sD_\tau$ meaning that $\sD_{\tau+1} = \sD_\tau \cup \{\vtheta_\tau\}$. Let $\boldsymbol{\rho}_j$ denotes the $j$-th model parameter in $\sD_t$. It can be concluded that $\boldsymbol{\rho}_j = \vtheta_{(j-1)n + 1}$. The server continues saving model parameters every $n$ time steps until time step $U \le T$. Client $i$ assigns a weight $w_{ij,t}$ to the $j$-th model in $\sD_t$, which assesses the credibility of the predictions given by the model parameter $\boldsymbol{\rho}_j$. Client $i$ initializes $w_{ij,1}=1$. At each time step $t$, client $i$ selects $M$ models with replacement from $\sD_t$ to construct its model set $\sM_{i,t}$. \Algref{alg:2} illustrates the model selection process conducted by client $i$ at time step $t$. During each round of model selection, client $i$ chooses a model according to a probability mass function (PMF) proportional to weights $\{w_{ij,t}|1 \le j \le |\sD_t|\}$ where $|\cdot|$ denotes the cardinality of a set (see step \ref{step:31} in \Algref{alg:2}). Client $i$ adds the chosen model to $\sM_{i,t}$ if it is not already present. Therefore, it can be concluded that $|\sM_{i,t}|\le M$, $\forall i, t$. Then at each time step $t$, client $i$ downloads all models in $\sM_{i,t}$ from the server.
Upon receiving the models, client $i$ constructs an ensemble model $\Tilde{f}_{i,t}(\cdot)$ as
\begin{align}
    \Tilde{f}_{i,t}(\vx) = \sum_{j \in \sM_{i,t}}{\frac{w_{ij,t}}{\sum_{m \in \sM_{i,t}}{w_{im,t}}}f(\vx;\boldsymbol{\rho}_j)}. \label{eq:20}
\end{align}
Client $i$ makes the prediction for $\vx_{i,t}$ as
\begin{align}
    \bar{f}_{i,t}(\vx_{i,t}) = \frac{\gamma_{i,t}}{\gamma_{i,t}+\delta_{i,t}}f_{i,t}(\vx_{i,t})+\frac{\delta_{i,t}}{\gamma_{i,t}+\delta_{i,t}}\Tilde{f}_{i,t}(\vx_{i,t}) \label{eq:14}
\end{align}
where $f_{i,t}(\vx_{i,t})$ is the ensemble of local and federated models as defined in \eqref{eq:7}. Furthermore, $\gamma_{i,t}$ and $\delta_{i,t}$ are weights assigned by client $i$ to ensemble models $f_{i,t}(\cdot)$ and $\Tilde{f}_{i,t}(\cdot)$, respectively. Upon observing the label $y_{i,t}$ after prediction, client $i$ updates weights $\gamma_{i,t}$ and $\delta_{i,t}$ as
\begin{subequations} \label{eq:18}
\begin{align}
    \gamma_{i,t+1} &= \gamma_{i,t}\exp\left(-\eta_c \gL(f_{i,t}(\vx_{i,t}),y_{i,t})\right), \label{eq:18a} \\
    \delta_{i,t+1} &= \delta_{i,t}\exp(-\eta_c \gL(\Tilde{f}_{i,t}(\vx_{i,t}),y_{i,t})). \label{eq:18b}
\end{align}
\end{subequations}
Furthermore, $\alpha_{i,t}$ and $\beta_{i,t}$ which are used to construct $f_{i,t}(\vx_{i,t})$ in \eqref{eq:7} are updated as in \eqref{eq:8}. The weight $w_{ij,t}$ is updated using the importance sampling loss as
\begin{align}
    w_{ij,t+1} = w_{ij,t}\exp\left(-\eta_c\frac{\gL(f(\vx_{i,t};\boldsymbol{\rho}_j),y_{i,t})}{q_{ij,t}}\1_{j\in\sM_{i,t}}\right) \label{eq:15}
\end{align}
where $\1_{j\in\sM_{i,t}}$ denotes the indicator function and is 1 if $j\in\sM_{i,t}$. Moreover, $q_{ij,t}$ is the probability that client $i$ selects the $j$-th model in $\sD_t$ to be in $\sM_{i,t}$ and can be expressed as $q_{ij,t} = 1-(1-p_{ij,t})^M$ where $p_{ij,t}$ is defined in step \ref{step:31} of \Algref{alg:2}. The proposed algorithm, named Federated Learning with Personalized Online Ensemble (Fed-POE) is summarized in \Algref{alg:1}.
\begin{algorithm}[tb]
	\caption{Fed-POE: Federated Learning with Personalized Online Ensemble}
	\label{alg:1}
	\begin{algorithmic}[1]
		\STATE {\bfseries Input:}{ Model $f(\cdot;\cdot)$, batch size $b$, $n$, $U$, $\sD_{i,1}=\varnothing$.}
		\FOR{$t=1,\ldots,T$}
            \STATE The server transmits $\vtheta_t$ to all clients.
		\FORALL{$i \in [N]$, client $i$}
            \STATE Performs model selection given $\sD_t$ according to \Algref{alg:2} to obtain $\sM_{i,t}$.
            \STATE Makes prediction $\bar{f}_{i,t}(\vx_{i,t})$ as in \eqref{eq:14} using chosen model set $\sM_{i,t}$.
            \STATE Upon receiving $y_{i,t}$, updates $\alpha_{i,t}$, $\beta_{i,t}$, $\gamma_{i,t}$, $\delta_{i,t}$ and $\{w_{ij,t}\}_{j=1}^{|\sD_t|}$ as in \eqref{eq:8}, \eqref{eq:18} and \eqref{eq:15}.
            \STATE Updates the local model as $\boldsymbol{\phi}_{i,t+1} = \boldsymbol{\phi}_{i,t} - \frac{\eta}{b} \sum_{\tau=t-b}^{t}{\nabla \gL(f(\vx_{i,t};\boldsymbol{\phi}_{i,\tau}),y_{i,t})}$.
            \STATE Updates the federated model as $\boldsymbol{\psi}_{i,t+1} = \vtheta_t - \frac{\eta}{b} \sum_{\tau=t-b}^{t}{\nabla \gL(f(\vx_{i,\tau};\vtheta_t),y_{i,\tau})}$. \label{step:8}
            \STATE Sends $\boldsymbol{\psi}_{i,t+1}$ to the server.
            \ENDFOR
            \IF{$t\le U$ and $t \bmod n = 0$}
            \STATE The server updates $\sD_t$ as $\sD_{t+1}=\sD_t \cup \{\vtheta_t\}$.
            \ENDIF
            \STATE The server updates $\vtheta_t$ as $\vtheta_{t+1} = \frac{1}{N}\sum_{i=1}^{N}{\boldsymbol{\psi}_{i,t+1}}$. \label{step:10}
		\ENDFOR
	\end{algorithmic}
\end{algorithm}
It is useful to note that the proposed method in Subsection \ref{ens} is a special case of Fed-POE by setting $M=0$ and $b=1$.

\textbf{Efficiency of Fed-POE.} To perform model selection using Fed-POE, clients do not need to store all model parameters in $\sD_t$ for all $t \in [T]$. Instead, the server, which has higher storage capacity than the clients, stores all model parameters, and clients download a subset of at most $M$ model parameters. The hyperparameter $M$ can be chosen such that clients can handle the memory and computational requirements of making predictions with the selected subset of models. Let $C_F$ denote the number of computations required to fine-tune the model $f$, and let $C_I$ represent the number of computations required to make an inference with model $f$. Assume that the complexity of model selection in Algorithm 1 is negligible compared to fine-tuning and making inferences with model $f$. According to Algorithm 2, each client performs $2C_F + (M+2)C_I$ computations per time step. Therefore, the computational complexity of Fed-POE for each client is $\mathcal{O}(C_F + MC_I)$. Beyond memory and computational considerations, selecting a subset of models from $\sD_t$ helps clients improve their prediction accuracy. Specifically, using the model weights $\{w_{ij,t}\}$ at time step $t$, client $i$ selects models that perform better on its data while pruning those with lower performance.

Let $h_j(\vx_{i,t})=f(\vx_{i,t};\boldsymbol{\rho}_j)$ denote the model associated with the $j$-th model parameter in $\sD_t$. Furthermore, %$h_{\text{loc}}(\cdot)$ and $h_{\text{fed}}(\cdot)$ defined as
$h_{\text{loc}}(\vx_{i,t})=f(\vx_{i,t};\boldsymbol{\phi}_{i,t})$ and $h_{\text{fed}}(\vx_{i,t})=f(\vx_{i,t};\vtheta_t)$ represent the local and federated models, respectively, . Let the set of models $\sH$ be defined as $\sH:=\{h_j \mid \forall j: 1 \le j \le |\sD_T|\} \cup \{h_{\text{loc}}, h_{\text{fed}}\}$. This set $\sH$ includes all models that can be employed by each client using Fed-POE. The best model in hindsight $h^*$ and the best model in hindsight for client $i$, $h_i^*$ are defined as
\begin{subequations} \label{eq:16}
    \begin{align}
        h^* &= \min_{h\in \sH}{\sum_{t=1}^{T}{\sum_{i=1}^{N}{\gL(h(\vx_{i,t}),y_{i,t})}}}, \label{eq:16a} \\
        h_i^* &= \min_{h\in \sH}{\sum_{t=1}^{T}{\gL(h(\vx_{i,t}),y_{i,t})}}. \label{eq:16b}
    \end{align}
\end{subequations}
The following theorem provides the regret upper bound of Fed-POE.

\begin{theorem} \label{th:3}
     Under assumption A3, employing Fed-POE in \Algref{alg:1}, the expected global regret of all clients is bounded from above as 
     \begin{align}
         &\frac{1}{N}\sum_{t=1}^{T}{\sum_{i=1}^{N}{\E_t[\gL(\bar{f}_{i,t}(\vx_{i,t}),y_{i,t})]}} - \frac{1}{N}\sum_{t=1}^{T}{\sum_{i=1}^{N}{\gL(h^*(\vx_{i,t}),y_{i,t})}} \nonumber \\ \le& \frac{\ln 2U - \ln 2n}{\eta_c} + \frac{\eta_c}{2} (\frac{U}{n}+1)T + (1-\frac{\eta_c}{2n}U)U  \label{eq:17}
     \end{align}
     while client $i$ achieves the following expected personalized regret upper bound:
     \begin{align}
         &\sum_{t=1}^{T}{\E_t[\gL(\bar{f}_{i,t}(\vx_{i,t}),y_{i,t})]} - \sum_{t=1}^{T}{\gL(h_i^*(\vx_{i,t}),y_{i,t})} \nonumber \\ \le& \frac{\ln 2U - \ln 2n}{\eta_c} + \frac{\eta_c}{2} (\frac{U}{n}+1)T + (1-\frac{\eta_c}{2n}U)U.  \label{eq:21}
     \end{align}
     The expectation is taken with respect to randomization in model selection.
\end{theorem}
The proof of Theorem \ref{th:3} can be found in Appendix \ref{C}. According to \eqref{eq:17} and \eqref{eq:21}, setting $\eta_c = \gO(1/\sqrt{T})$, $n=\gO(\sqrt{T})$ and $U=\gO(\sqrt{T})$, both the personalized and global regrets of clients achieve sublinear regret of $\gO(\sqrt{T})$. Since clients construct their ensemble models using a time-varying subset of models, employing existing model selection and ensemble learning approaches \cite{Fern2003,Foster2017,Muthukumar2019,Karimi2021,Ghari2022EUSIPCO,Pacchiano2022,Ghari2024} may not ensure the sublinear regrets stated in Theorem \ref{th:3}. However, by using the proposed \Algref{alg:2}, Fed-POE guarantees sublinear regret bounds while allowing clients the flexibility to select time-varying and personalized subsets of models for their ensembles.

\section{Experiments} \label{sec:exp}
The present section studies the performance of Fed-POE in \Algref{alg:1} compared to other baselines. Experiments are conducted on both image classification and regression tasks. The performance of federated learning is examined in both convex and non-convex cases. Codes are available at \url{https://github.com/pouyamghari/Fed-POE}. %Specifically, for the image classification task, clients and the server collaborate to fine-tune a pre-trained convolutional neural network (CNN) while for the regression, clients and the server collaborate to train a random-feature-based kernel model which is a convex model.

% \begin{table}[t]
\begin{wraptable}{R}{.5\textwidth} 
\vspace{-.8cm}
\caption{Average MSE ($\times 10^{-3}$) and its standard deviation ($\times 10^{-3}$) across clients for online regression on Air and WEC datasets.}
\label{table:3}
\begin{center}
\begin{tabular}{l||c|c}
\toprule
{\bf Methods}   &Air  &WEC 
\\ \hline 
Local   &$9.12 \pm 3.59$   &$17.64 \pm 0.44$ \\
Ditto   &$10.65 \pm 5.69$   &$33.88 \pm 16.08$ \\
Fed-Rep   &$10.48 \pm 5.23$   &$35.13 \pm 10.38$ \\
Fed-OMD   &$11.48 \pm 6.84$   &$32.61 \pm 27.38$ \\
eM-KOFL   &$11.51 \pm 6.71$   &$72.29 \pm 62.48$ \\
POF-MKL   &$10.66 \pm 6.07$   &$16.94 \pm 15.92$ \\
\hline
Fed-POE   &$\mathbf{9.06} \pm \mathbf{3.73}$   &$\mathbf{11.83} \pm \mathbf{4.60}$ \\
\bottomrule
\end{tabular}
\end{center}
\vspace{-.2cm}
% \end{table}
\end{wraptable}

\subsection{Regression} \label{exp:reg}
The performance of the proposed Fed-POE is evaluated on online regression tasks. For these tasks, clients and the server collaborate to train a random feature kernel-based model, which is known to be convex \cite{Hoi2013,Sahoo2014,Ghari2020}. Details about the random feature kernel-based model used in the experiments can be found in Appendix \ref{D}. The performance of Fed-POE is compared with a baseline called Local, where clients train their models locally without participating in federated learning. Additionally, Fed-POE is compared to personalized federated learning baselines Ditto \cite{TLi2021} and Fed-Rep \cite{Collins2021}, the online federated learning baseline Fed-OMD \cite{Mitra2021}, and online federated kernel learning baselines eM-KOFL \cite{Hong2022} and POF-MKL \cite{Ghari2022}.
%Both vM-KOFL and POF-MKL are multi-kernel learning algorithms. In order to construct a random feature kernel-based model, three Gaussian kernels with variances of $0.1$, $1$ and $10$ are employed. In order to implement Fed-POE with multiple kernels, at each time step each client makes a prediction by linearly combining predictions from $3$ Gaussian kernels. Then each client locally updates the weight for each Gaussian kernel via multiplicative update rule (see e.g. \cite{Hazan2022}). For all online federated kernel learning algorithms, the number of random features is set to $D=100$ for each kernel and $20$ different sets of random features are sampled. 
Mean square error (MSE) is used as the metric to evaluate the performance of algorithms on regression task. MSE for client $i$ can be expressed as
$
    \text{MSE}_i = {\frac{1}{T}\sum_{t=1}^{T}{(\hat{y}_{ij,t} - y_{i,t})^2}} %\label{eq:19}
$
where $\hat{y}_{ij,t}$ denotes the prediction of client $i$ at time step $t$. The performance of algorithms are tested on two regression datasets Air \cite{Zhang2017} and WEC \cite{Neshat2018}. Air dataset is a time-series dataset that each data sample contains air quality features and the goal is to predict the concentration of contamination in the air. Each sample in WEC dataset contains features of different wave energy converters and the goal is to predict power output. Data samples are distributed non-i.i.d among $400$ clients. Time horizon $T$ for both datasets is $250$. 
More information about datasets and distributed data among clients is presented in Appendix \ref{D}.

Table \ref{table:3} presents the MSE of algorithms and their standard deviation across clients. For all algorithms, the learning rates are set to $\eta=\eta_c=1/\sqrt{T}$. Table \ref{table:3} shows that when data is distributed non-i.i.d. among clients, local model training can achieve higher accuracy compared to federated learning. For the Air dataset, Local achieves lower MSE than other federated learning baselines except for Fed-POE. For the WEC dataset, only POF-MKL achieves lower MSE than Local. This indicates that the performance of federated learning approaches compared to Local depends on the dataset. By utilizing both federated and local models, Fed-POE achieves the lowest MSE. Table \ref{table:3} shows that the performance of Fed-POE relative to other baselines is more consistent across different datasets.

% \begin{table}[t]
% % \setlength{\tabcolsep}{4pt}
% \caption{Average MSE ($\times 10^{-3}$) and its standard deviation ($\times 10^{-3}$) across clients for online regression on Air and WEC datasets. The value of $NT$ is fixed as $NT=100,000$}
% \label{table:3}
% \begin{center}
% \begin{tabular}{l||c|c|c|c}
% \toprule
% {\bf Methods}    &Air~($N=100$)    &Air~($N=400$) &WEC~($N=100)$  &WEC~($N=400$) 
% \\ \hline 
% Local  &$\mathbf{7.66} \pm \mathbf{2.36}$   &$\mathbf{9.12} \pm \mathbf{3.59}$   &$8.64 \pm 0.10$   &$17.64 \pm 0.44$ \\
% Fed-OMD  &$11.45 \pm 3.91$   &$12.50 \pm 5.02$  &$11.45 \pm 3.91$   &$12.50 \pm 5.02$ \\
% vM-KOFL  &$10.02 \pm 3.98$   &$11.48 \pm 6.84$  &$16.83 \pm 14.03$   &$32.61 \pm 2.74$ \\
% POF-MKL  &$8.67 \pm 3.27$   &$10.66 \pm 6.07$   &$8.47 \pm 7.68$   &$16.94 \pm 15.92$ \\
% \hline
% Fed-POE  &$\mathbf{7.68} \pm \mathbf{2.47}$   &$\mathbf{9.06} \pm \mathbf{3.73}$  &$\mathbf{5.86} \pm \mathbf{2.12}$   &$\mathbf{11.83} \pm \mathbf{4.60}$ \\
% \bottomrule
% \end{tabular}
% \end{center}
% \end{table}

\subsection{Image Classification}
The performance of the proposed Fed-POE on an image classification task is compared with Local, Ditto \cite{TLi2021}, Fed-Rep \cite{Collins2021}, Fed-OMD \cite{Mitra2021}, Fed-ALA \cite{Zhang2023}, and Fed-DS \cite{Marfoq2023}. Fed-ALA \cite{Zhang2023} is a personalized federated learning model suitable for deep neural networks, while Fed-DS \cite{Marfoq2023} is a federated learning algorithm designed to handle data streams. %In order to implement Fed-OMD, $\ell_2$-norm is employed as the regularization function. 
Experiments are conducted on CIFAR-10 \cite{Krizhevsky2009} and Fashion MNIST (FMNIST) \cite{Xiao2017} datasets. CIFAR-10 and FMNIST contain $60,000$ and $70,000$ images. For both CIFAR-10 and FMNIST, a CNN with a VGG architecture \cite{Simonyan2015}, consisting of two blocks, is pre-trained on a subset of training samples from each dataset. The training datasets are biased towards class $0$. More details about training the CNNs can be found in Appendix \ref{D}. %Given a pre-trained CNN, clients and the server employ online federated learning algorithms to both perform image classification and fine-tune the last layer of the pre-trained CNN in an online fashion. At each time step $t$, the server sends the fine-tuned CNN to clients. At time step $t$, each client receives an image and performs classification on the received image. Then the label of the image is revealed to the client and the client fine-tunes the last layer of the CNN locally. Clients send their locally fine-tuned CNNs to the server. Aggregating local updates, the server fine-tunes the last layer of the CNN globally and send the global model to clients to be used in the next time step. This continues up until the time horizon $T$. 
For both the CIFAR-10 and FMNIST datasets, $10,000$ test samples are sequentially received by clients. There are $20$ clients in total, and the data samples are distributed non-i.i.d. among them. For CIFAR-10, each client is biased towards one specific class, with $55\%$ of the samples belonging to that class and $5\%$ of the samples belonging to each of the other 9 classes. For FMNIST, each client is biased towards two classes, and the distribution of samples is time-variant. More details about the data distribution among clients and experimental setup can be found in Appendix \ref{D}. At each time step, all algorithms employ batch of size $10$ for model update. The learning rates for all algorithms are set to $\eta = 0.01/\sqrt{T}$ and $\eta_c=1/\sqrt{T}$. The server stores models every $n=20$ time steps. The metric to evaluate the performance of algorithms is the accuracy. The accuracy for client $i$ is defined as
$
    \text{Accuracy}_i = \frac{1}{T} \sum_{t=1}^{T}{\1_{\hat{y}_{i,t} = y_{i,t}}}
$
where $\hat{y}_{i,t}$ denotes the label predicted by client $i$ at time step $t$.
% \begin{table}[t]
\begin{wraptable}{R}{.52\textwidth}
\vspace{-.5cm}
\setlength{\tabcolsep}{2pt}
\caption{Average accuracy and its standard deviation across clients for image classification.}
\label{table:1}
\begin{center}
\begin{tabular}{l||c|c}
\toprule
{\bf Methods}    &CIFAR-10   &FMNIST 
\\ \hline 
Local  &$50.35\% \pm 10.11\%$   &$78.81\% \pm 2.12\%$ \\
Ditto  &$56.87\% \pm 9.06\%$   &$78.73\% \pm 1.89\%$ \\
Fed-Rep  &$63.86\% \pm 7.97\%$   &$79.04\% \pm 1.77\%$ \\
Fed-OMD  &$65.09\% \pm 7.39\%$  &$74.60\% \pm 6.52\%$ \\
Fed-ALA  &$61.48\% \pm 8.88\%$   &$75.13\% \pm 6.39\%$ \\
Fed-DS  &$64.26\% \pm 7.03\%$   &$75.62\% \pm 6.58\%$ \\
\hline
Fed-POE  &$\mathbf{66.54}\% \pm \mathbf{8.07}\%$  &$\mathbf{79.23}\% \pm \mathbf{1.88}\%$ \\
\bottomrule
\end{tabular}
\end{center}
\vspace{-.4cm}
% \end{table}
\end{wraptable}

Average accuracy and its standard deviation across clients for CIFAR-10 and FMNIST are reported in Table \ref{table:1}. At each time step $t$, clients can store $10$ model parameters. Therefore, $M$ is set to $M=8$ for Fed-POE. The results indicate that the performance of Local relative to federated learning baselines depends on the dataset. While Local outperforms all federated baselines except for Fed-Rep and Fed-POE on FMNIST, it obtains the worst accuracy on CIFAR-10. Conversely, Fed-POE achieves the highest accuracy for both datasets, indicating that Fed-POE efficiently leverages the advantages of both federated and local models. %In summary, the results in Tables \ref{table:3} and \ref{table:1} demonstrate the efficacy of Fed-POE in online prediction and model fine-tuning.
\begin{table}[t]
\setlength{\tabcolsep}{4pt}
\caption{Average accuracy and standard deviation across clients employing Fed-POE for image classification on CIFAR-10 with varying values of $M$ and batch size $b$.}
\label{table:4}
\begin{center}
\begin{tabular}{l||c|c|c|c}
\toprule
    &$M=0$ &$M=4$ &$M=8$ &$M=16$
\\ \hline
$b=1$  &$53.80\% \pm 6.71\%$   &$62.73\% \pm 8.29\%$ &$62.73\% \pm 8.29\%$   &$62.73\% \pm 8.26\%$ \\
$b=10$  &$65.55\% \pm 8.77\%$   &$66.50\% \pm 8.00\%$ &$66.54\% \pm 8.08\%$   &$66.46\% \pm 7.98\%$ \\
$b=20$  &$65.72\% \pm 8.62\%$   &$66.13\% \pm 8.20\%$ &$66.64\% \pm 7.94\%$   &$66.53\% \pm 8.00\%$ \\
$b=30$  &$66.83\% \pm 8.54\%$   &$66.32\% \pm 7.92\%$ &$66.24\% \pm 8.05\%$   &$66.39\% \pm 8.02\%$ \\
\bottomrule
\end{tabular}
\end{center}
\end{table}
To analyze the effect of the number of models $M$ and batch size $b$ on the Fed-POE performance, experiments are conducted on the CIFAR-10 dataset, varying the batch size $b$ and the number of models $M$ selected by each client to construct the ensemble model. As observed in Table \ref{table:4}, the batch size $b=1$ results in the worst accuracy, mainly due to the forgetting process where models overfit to the most recently observed data. However, increasing the batch size from $b=10$ or $b=20$ to $b=30$ does not significantly improve the accuracy. Larger batch sizes may lead the model to perform better on older data, as the model is trained on older data over more iterations. Therefore, this study concludes that a moderate batch size is optimal, considering that increasing the batch size also increases computational complexity. Table \ref{table:2} in Appendix \ref{D} presents the accuracy of Fed-POE on both the CIFAR-10 and FMNIST datasets with varying values of $M$. The table shows that employing the saved models by the server in $\sD_t$ improves performance, as setting $M=0$ results in the worst accuracy. Moreover, it can be observed that increasing $M$ does not necessarily lead to further accuracy improvement. This aligns with the intuition behind selecting a subset of models rather than using all models.

\section{Conclusions}
This paper proposed Fed-POE, a personalized federated learning algorithm designed for online prediction and model fine-tuning. Fed-POE constructs an ensemble using local models and federated models stored by the server periodically over time. Theoretical analysis demonstrated that Fed-POE achieves sublinear regret. Experimental results revealed that the relative performance of local models compared to federated models depends on the dataset, making the decision between local model training and federated learning challenging. However, experimental results also show that Fed-POE consistently outperforms both local and federated models across all datasets. This indicates that Fed-POE effectively leverages the advantages of both local and federated models.

\section*{Acknowledgement}
Work in this paper is supported by NSF ECCS 2207457 and NSF ECCS 2412484.

\bibliographystyle{plainnat}
\bibliography{References}

\begin{thebibliography}{58}
\providecommand{\natexlab}[1]{#1}
\providecommand{\url}[1]{\texttt{#1}}
\expandafter\ifx\csname urlstyle\endcsname\relax
  \providecommand{\doi}[1]{doi: #1}\else
  \providecommand{\doi}{doi: \begingroup \urlstyle{rm}\Url}\fi

\bibitem[Acar et~al.(2021)Acar, Zhao, Zhu, Matas, Mattina, Whatmough, and
  Saligrama]{Acar2021}
Durmus Alp~Emre Acar, Yue Zhao, Ruizhao Zhu, Ramon Matas, Matthew Mattina, Paul
  Whatmough, and Venkatesh Saligrama.
\newblock Debiasing model updates for improving personalized federated
  training.
\newblock In \emph{Proceedings of International Conference on Machine
  Learning}, volume 139, pages 21--31, Jul 2021.

\bibitem[Alon et~al.(2017)Alon, Cesa-Bianchi, Gentile, Mannor, Mansour, and
  Shamir]{Alon2017}
Noga Alon, Nicolò Cesa-Bianchi, Claudio Gentile, Shie Mannor, Yishay Mansour,
  and Ohad Shamir.
\newblock Nonstochastic multi-armed bandits with graph-structured feedback.
\newblock \emph{SIAM Journal on Computing}, 46\penalty0 (6):\penalty0
  1785--1826, 2017.

\bibitem[Auer et~al.(2003)Auer, Cesa-Bianchi, Freund, and Schapire]{Auer2003}
Peter Auer, Nicol\`{o} Cesa-Bianchi, Yoav Freund, and Robert~E. Schapire.
\newblock The nonstochastic multiarmed bandit problem.
\newblock \emph{SIAM Journal on Computing}, 32\penalty0 (1):\penalty0 48–77,
  Jan 2003.

\bibitem[Cesa-Bianchi and Lugosi(2006)]{Cesa-Bianchi2006}
Nicolo Cesa-Bianchi and Gabor Lugosi.
\newblock \emph{Prediction, Learning, and Games}.
\newblock Cambridge University Press, USA, 2006.

\bibitem[Charles et~al.(2021)Charles, Garrett, Huo, Shmulyian, and
  Smith]{Charles2021}
Zachary Charles, Zachary Garrett, Zhouyuan Huo, Sergei Shmulyian, and Virginia
  Smith.
\newblock On large-cohort training for federated learning.
\newblock In \emph{Advances in Neural Information Processing Systems}, 2021.

\bibitem[Chen et~al.(2022)Chen, Ding, Tramel, Wu, Sahu, Avestimehr, and
  Zhang]{Chen2022}
Huili Chen, Jie Ding, Eric~W Tramel, Shuang Wu, Anit~Kumar Sahu, Salman
  Avestimehr, and Tao Zhang.
\newblock Self-aware personalized federated learning.
\newblock \emph{Advances in Neural Information Processing Systems},
  35:\penalty0 20675--20688, 2022.

\bibitem[Chen et~al.(2020)Chen, Ning, Slawski, and Rangwala]{Chen2020}
Yujing Chen, Yue Ning, Martin Slawski, and Huzefa Rangwala.
\newblock Asynchronous online federated learning for edge devices with non-iid
  data.
\newblock In \emph{IEEE International Conference on Big Data (Big Data)}, pages
  15--24, Dec 2020.

\bibitem[Collins et~al.(2021)Collins, Hassani, Mokhtari, and
  Shakkottai]{Collins2021}
Liam Collins, Hamed Hassani, Aryan Mokhtari, and Sanjay Shakkottai.
\newblock Exploiting shared representations for personalized federated
  learning.
\newblock In \emph{Proceedings of the International Conference on Machine
  Learning}, volume 139, pages 2089--2099, Jul 2021.

\bibitem[Damaskinos et~al.(2022)Damaskinos, Guerraoui, Kermarrec, Nitu, Patra,
  and Taiani]{Damaskinos2022}
Georgios Damaskinos, Rachid Guerraoui, Anne-Marie Kermarrec, Vlad Nitu,
  Rhicheek Patra, and Francois Taiani.
\newblock Fleet: Online federated learning via staleness awareness and
  performance prediction.
\newblock \emph{ACM Transactions on Intelligent Systems and Technology},
  13\penalty0 (5), Sep 2022.

\bibitem[Deng et~al.(2020)Deng, Kamani, and Mahdavi]{Deng2020}
Yuyang Deng, Mohammad~Mahdi Kamani, and Mehrdad Mahdavi.
\newblock Adaptive personalized federated learning.
\newblock \emph{arXiv preprint arXiv:2003.13461}, 2020.

\bibitem[Dinh et~al.(2020)Dinh, Tran, and Nguyen]{Dinh2020}
Canh~T. Dinh, Nguyen~H. Tran, and Tuan~Dung Nguyen.
\newblock Personalized federated learning with moreau envelopes.
\newblock In \emph{Proceedings of International Conference on Neural
  Information Processing Systems}, page 21394–21405, Dec 2020.

\bibitem[Fallah et~al.(2020)Fallah, Mokhtari, and Ozdaglar]{Fallah2020}
Alireza Fallah, Aryan Mokhtari, and Asuman Ozdaglar.
\newblock Personalized federated learning with theoretical guarantees: A
  model-agnostic meta-learning approach.
\newblock In \emph{Advances in Neural Information Processing Systems},
  volume~33, pages 3557--3568, Dec 2020.

\bibitem[Fern and Givan(2003)]{Fern2003}
Alan Fern and Robert Givan.
\newblock Online ensemble learning: An empirical study.
\newblock \emph{Machine Learning}, 53:\penalty0 71--109, 2003.

\bibitem[Finn et~al.(2017)Finn, Abbeel, and Levine]{Finn2017}
Chelsea Finn, Pieter Abbeel, and Sergey Levine.
\newblock Model-agnostic meta-learning for fast adaptation of deep networks.
\newblock In \emph{Proceedings of International Conference on Machine
  Learning}, volume~70, pages 1126--1135, Aug 2017.

\bibitem[Foster et~al.(2017)Foster, Kale, Mohri, and Sridharan]{Foster2017}
Dylan Foster, Satyen Kale, Mehryar Mohri, and Karthik Sridharan.
\newblock Parameter-free online learning via model selection.
\newblock In \emph{Proceedings of International Conference on Neural
  Information Processing Systems}, 2017.

\bibitem[Ganguly and Aggarwal(2022)]{Ganguly2022}
Bhargav Ganguly and Vaneet Aggarwal.
\newblock Online federated learning via non-stationary detection and adaptation
  amidst concept drift.
\newblock \emph{arXiv preprint arXiv:2211.12578}, 2022.

\bibitem[Gauthier et~al.(2022)Gauthier, Gogineni, Werner, Huang, and
  Kuh]{Gauthier2022}
Francois Gauthier, Vinay~Chakravarthi Gogineni, Stefan Werner, Yih-Fang Huang,
  and Anthony Kuh.
\newblock Resource-aware asynchronous online federated learning for nonlinear
  regression.
\newblock In \emph{IEEE International Conference on Communications}, pages
  2828--2833, 2022.

\bibitem[Ghari and Shen(2020)]{Ghari2020}
Pouya~M Ghari and Yanning Shen.
\newblock Online multi-kernel learning with graph-structured feedback.
\newblock In \emph{Proceedings of the International Conference on Machine
  Learning}, volume 119, pages 3474--3483, Jul 2020.

\bibitem[Ghari and Shen(2022{\natexlab{a}})]{Ghari2022}
Pouya~M. Ghari and Yanning Shen.
\newblock Personalized online federated learning with multiple kernels.
\newblock In \emph{Advances in Neural Information Processing Systems},
  2022{\natexlab{a}}.

\bibitem[Ghari and Shen(2022{\natexlab{b}})]{Ghari2022EUSIPCO}
Pouya~M. Ghari and Yanning Shen.
\newblock Graph-assisted communication-efficient ensemble federated learning.
\newblock In \emph{European Signal Processing Conference (EUSIPCO)}, pages
  737--741, 2022{\natexlab{b}}.

\bibitem[Ghari and Shen(2024{\natexlab{a}})]{Ghari2024}
Pouya~M. Ghari and Yanning Shen.
\newblock Budgeted online model selection and fine-tuning via federated
  learning.
\newblock \emph{Transactions on Machine Learning Research}, 2024{\natexlab{a}}.

\bibitem[Ghari and Shen(2024{\natexlab{b}})]{Ghari2024tnnls}
Pouya~M. Ghari and Yanning Shen.
\newblock Online learning with uncertain feedback graphs.
\newblock \emph{IEEE Transactions on Neural Networks and Learning Systems},
  35\penalty0 (7):\penalty0 9636--9650, 2024{\natexlab{b}}.

\bibitem[Gogineni et~al.(2023)Gogineni, Werner, Huang, and Kuh]{Gogineni2023}
Vinay~Chakravarthi Gogineni, Stefan Werner, Yih-Fang Huang, and Anthony Kuh.
\newblock Communication-efficient online federated learning strategies for
  kernel regression.
\newblock \emph{IEEE Internet of Things Journal}, 10\penalty0 (5):\penalty0
  4531--4544, 2023.

\bibitem[Hanzely et~al.(2020)Hanzely, Hanzely, Horv\'{a}th, and
  Richt\'{a}rik]{Hanzely2020}
Filip Hanzely, Slavom\'{\i}r Hanzely, Samuel Horv\'{a}th, and Peter
  Richt\'{a}rik.
\newblock Lower bounds and optimal algorithms for personalized federated
  learning.
\newblock In \emph{Proceedings of International Conference on Neural
  Information Processing Systems}, page 2304–2315, Dec 2020.

\bibitem[Hoi et~al.(2013)Hoi, Jin, Zhao, and Yang]{Hoi2013}
Steven C.~H. Hoi, Rong Jin, Peilin Zhao, and Tianbao Yang.
\newblock Online multiple kernel classification.
\newblock \emph{Machine Learning}, 90:\penalty0 289–316, Feb 2013.

\bibitem[Hong and Chae(2022)]{Hong2022}
Songnam Hong and Jeongmin Chae.
\newblock Communication-efficient randomized algorithm for multi-kernel online
  federated learning.
\newblock \emph{IEEE Transactions on Pattern Analysis and Machine
  Intelligence}, 44\penalty0 (12):\penalty0 9872--9886, 2022.

\bibitem[Kelly et~al.(2023)Kelly, Longjohn, and Nottingham]{Kelly2023}
Markelle Kelly, Rachel Longjohn, and Kolby Nottingham.
\newblock {UCI} machine learning repository, 2023.

\bibitem[Krizhevsky(2009)]{Krizhevsky2009}
Alex Krizhevsky.
\newblock Learning multiple layers of features from tiny images.
\newblock Technical report, 2009.

\bibitem[Li et~al.(2020{\natexlab{a}})Li, Sahu, Talwalkar, and Smith]{Li2020}
Tian Li, Anit~Kumar Sahu, Ameet Talwalkar, and Virginia Smith.
\newblock Federated learning: Challenges, methods, and future directions.
\newblock \emph{IEEE Signal Processing Magazine}, 37\penalty0 (3):\penalty0
  50--60, 2020{\natexlab{a}}.

\bibitem[Li et~al.(2021{\natexlab{a}})Li, Hu, Beirami, and Smith]{TLi2021}
Tian Li, Shengyuan Hu, Ahmad Beirami, and Virginia Smith.
\newblock Ditto: Fair and robust federated learning through personalization.
\newblock In \emph{Proceedings of the International Conference on Machine
  Learning}, volume 139, pages 6357--6368, Jul 2021{\natexlab{a}}.

\bibitem[Li et~al.(2020{\natexlab{b}})Li, Huang, Yang, Wang, and
  Zhang]{Li2020b}
Xiang Li, Kaixuan Huang, Wenhao Yang, Shusen Wang, and Zhihua Zhang.
\newblock On the convergence of fedavg on non-iid data.
\newblock In \emph{International Conference on Learning Representations},
  2020{\natexlab{b}}.

\bibitem[Li et~al.(2021{\natexlab{b}})Li, Zhan, Shao, Li, and Song]{LiX2021}
Xin-Chun Li, De-Chuan Zhan, Yunfeng Shao, Bingshuai Li, and Shaoming Song.
\newblock Fedphp: Federated personalization with inherited private models.
\newblock In \emph{Joint European Conference on Machine Learning and Knowledge
  Discovery in Databases}, pages 587--602, 2021{\natexlab{b}}.

\bibitem[Liu et~al.(2022)Liu, Hu, Wu, and Smith]{Liu2022}
Ken Liu, Shengyuan Hu, Steven Wu, and Virginia Smith.
\newblock On privacy and personalization in cross-silo federated learning.
\newblock In \emph{Advances in Neural Information Processing Systems}, 2022.

\bibitem[Marfoq et~al.(2021)Marfoq, Neglia, Bellet, Kameni, and
  Vidal]{Marfoq2021}
Othmane Marfoq, Giovanni Neglia, Aur{\'e}lien Bellet, Laetitia Kameni, and
  Richard Vidal.
\newblock Federated multi-task learning under a mixture of distributions.
\newblock In \emph{Proceedings of International Conference on Neural
  Information Processing Systems}, volume~34, pages 15434--15447, Dec 2021.

\bibitem[Marfoq et~al.(2023)Marfoq, Neglia, Kameni, and Vidal]{Marfoq2023}
Othmane Marfoq, Giovanni Neglia, Laetitia Kameni, and Richard Vidal.
\newblock Federated learning for data streams.
\newblock In \emph{Proceedings of The International Conference on Artificial
  Intelligence and Statistics}, volume 206, pages 8889--8924, Apr 2023.

\bibitem[Mawuli et~al.(2023)Mawuli, Kumar, Nanor, Fu, Pan, Yang, Zhang, and
  Shao]{Mawuli2023}
Cobbinah~B. Mawuli, Jay Kumar, Ebenezer Nanor, Shangxuan Fu, Liangxu Pan, Qinli
  Yang, Wei Zhang, and Junming Shao.
\newblock Semi-supervised federated learning on evolving data streams.
\newblock \emph{Information Sciences}, 643:\penalty0 119235, 2023.

\bibitem[Mitra et~al.(2021)Mitra, Hassani, and Pappas]{Mitra2021}
Aritra Mitra, Hamed Hassani, and George~J. Pappas.
\newblock Online federated learning.
\newblock In \emph{IEEE Conference on Decision and Control (CDC)}, pages
  4083--4090, Dec 2021.

\bibitem[Mohri et~al.(2019)Mohri, Sivek, and Suresh]{Mohri2019}
Mehryar Mohri, Gary Sivek, and Ananda~Theertha Suresh.
\newblock Agnostic federated learning.
\newblock In Kamalika Chaudhuri and Ruslan Salakhutdinov, editors,
  \emph{Proceedings of the International Conference on Machine Learning},
  volume~97, pages 4615--4625, Jun 2019.

\bibitem[Muthukumar et~al.(2019)Muthukumar, Ray, Sahai, and
  Bartlett]{Muthukumar2019}
Vidya Muthukumar, Mitas Ray, Anant Sahai, and Peter Bartlett.
\newblock Best of many worlds: Robust model selection for online supervised
  learning.
\newblock In \emph{Proceedings of the International Conference on Artificial
  Intelligence and Statistics}, volume~89, pages 3177--3186, Apr 2019.

\bibitem[Neshat et~al.(2018)Neshat, Alexander, Wagner, and Xia]{Neshat2018}
Mehdi Neshat, Bradley Alexander, Markus Wagner, and Yuanzhong Xia.
\newblock A detailed comparison of meta-heuristic methods for optimising wave
  energy converter placements.
\newblock In \emph{Proceedings of the Genetic and Evolutionary Computation
  Conference}, page 1318–1325, Jul 2018.

\bibitem[Pacchiano et~al.(2022)Pacchiano, Dann, and Gentile]{Pacchiano2022}
Aldo Pacchiano, Christoph Dann, and Claudio Gentile.
\newblock Best of both worlds model selection.
\newblock In Alice~H. Oh, Alekh Agarwal, Danielle Belgrave, and Kyunghyun Cho,
  editors, \emph{Advances in Neural Information Processing Systems}, 2022.

\bibitem[Park et~al.(2022)Park, Kwon, and Hong]{Park2022}
Jonghwan Park, Dohyeok Kwon, and Songnam Hong.
\newblock Ofedqit: Communication-efficient online federated learning via
  quantization and intermittent transmission.
\newblock \emph{arXiv preprint arXiv:2205.06491}, 2022.

\bibitem[Patel et~al.(2023)Patel, Wang, Saha, and Srebro]{Patel2023}
Kumar~Kshitij Patel, Lingxiao Wang, Aadirupa Saha, and Nathan Srebro.
\newblock Federated online and bandit convex optimization.
\newblock In \emph{Proceedings of the International Conference on Machine
  Learning}, volume 202, pages 27439--27460, Jul 2023.

\bibitem[Rahimi and Recht(2007)]{Rahimi2007}
Ali Rahimi and Benjamin Recht.
\newblock Random features for large-scale kernel machines.
\newblock In \emph{Proceedings of International Conference on Neural
  Information Processing Systems}, pages 1177--1184, Dec 2007.

\bibitem[Ramasesh et~al.(2022)Ramasesh, Lewkowycz, and Dyer]{Ramasesh2022}
Vinay~Venkatesh Ramasesh, Aitor Lewkowycz, and Ethan Dyer.
\newblock Effect of scale on catastrophic forgetting in neural networks.
\newblock In \emph{International Conference on Learning Representations}, 2022.

\bibitem[Reza~Karimi et~al.(2021)Reza~Karimi, Merve~G{\"u}rel, Karla\v{s},
  Rausch, Zhang, and Krause]{Karimi2021}
Mohammad Reza~Karimi, Nezihe Merve~G{\"u}rel, Bojan Karla\v{s}, Johannes
  Rausch, Ce~Zhang, and Andreas Krause.
\newblock Online active model selection for pre-trained classifiers.
\newblock In \emph{Proceedings of The International Conference on Artificial
  Intelligence and Statistics}, volume 130, pages 307--315, Apr 2021.

\bibitem[Rothchild et~al.(2020)Rothchild, Panda, Ullah, Ivkin, Stoica,
  Braverman, Gonzalez, and Arora]{Rothchild2020}
Daniel Rothchild, Ashwinee Panda, Enayat Ullah, Nikita Ivkin, Ion Stoica,
  Vladimir Braverman, Joseph Gonzalez, and Raman Arora.
\newblock {F}etch{SGD}: Communication-efficient federated learning with
  sketching.
\newblock In \emph{Proceedings of the International Conference on Machine
  Learning}, volume 119, pages 8253--8265, Jul 2020.

\bibitem[Sahoo et~al.(2014)Sahoo, Hoi, and Li]{Sahoo2014}
Doyen Sahoo, Steven~C.H. Hoi, and Bin Li.
\newblock Online multiple kernel regression.
\newblock In \emph{Proceedings of ACM SIGKDD International Conference on
  Knowledge Discovery and Data Mining}, page 293–302, 2014.

\bibitem[Simonyan and Zisserman(2015)]{Simonyan2015}
Karen Simonyan and Andrew Zisserman.
\newblock Very deep convolutional networks for large-scale image recognition.
\newblock In \emph{International Conference on Learning Representations}, May
  2015.
\newblock URL \url{http://arxiv.org/abs/1409.1556}.

\bibitem[Tan et~al.(2022)Tan, Long, Ma, Liu, Zhou, and Jiang]{Tan2022}
Yue Tan, Guodong Long, Jie Ma, Lu~Liu, Tianyi Zhou, and Jing Jiang.
\newblock Federated learning from pre-trained models: A contrastive learning
  approach.
\newblock \emph{Advances in neural information processing systems},
  35:\penalty0 19332--19344, 2022.

\bibitem[Toneva et~al.(2019)Toneva, Sordoni, Combes, Trischler, Bengio, and
  Gordon]{Toneva2019}
Mariya Toneva, Alessandro Sordoni, Remi Tachet~des Combes, Adam Trischler,
  Yoshua Bengio, and Geoffrey~J Gordon.
\newblock An empirical study of example forgetting during deep neural network
  learning.
\newblock In \emph{International Conference on Learning Representations}, 2019.

\bibitem[Wahba(1990)]{Wahba1990}
Grace Wahba.
\newblock \emph{Spline Models for Observational Data}.
\newblock Society for Industrial and Applied Mathematics, 1990.

\bibitem[Wang et~al.(2022)Wang, Lin, and Chen]{Wang2022}
Yujia Wang, Lu~Lin, and Jinghui Chen.
\newblock Communication-efficient adaptive federated learning.
\newblock In \emph{Proceedings of the International Conference on Machine
  Learning}, volume 162, pages 22802--22838, Jul 2022.

\bibitem[Wu et~al.(2023)Wu, Zhang, Yu, Liu, Gu, Zhou, Chen, and Cheng]{Wu2023}
Yue Wu, Shuaicheng Zhang, Wenchao Yu, Yanchi Liu, Quanquan Gu, Dawei Zhou,
  Haifeng Chen, and Wei Cheng.
\newblock Personalized federated learning under mixture of distributions.
\newblock In \emph{International Conference on Machine Learning}, pages
  37860--37879, 2023.

\bibitem[Xiao et~al.(2017)Xiao, Rasul, and Vollgraf]{Xiao2017}
Han Xiao, Kashif Rasul, and Roland Vollgraf.
\newblock Fashion-mnist: a novel image dataset for benchmarking machine
  learning algorithms.
\newblock \emph{arXiv preprint arXiv:1708.07747}, 2017.

\bibitem[Ye et~al.(2023)Ye, Ni, Wu, Chen, and Wang]{Ye2023}
Rui Ye, Zhenyang Ni, Fangzhao Wu, Siheng Chen, and Yanfeng Wang.
\newblock Personalized federated learning with inferred collaboration graphs.
\newblock In \emph{International Conference on Machine Learning}, pages
  39801--39817, 2023.

\bibitem[Zhang et~al.(2023)Zhang, Hua, Wang, Song, Xue, Ma, and
  Guan]{Zhang2023}
Jianqing Zhang, Yang Hua, Hao Wang, Tao Song, Zhengui Xue, Ruhui Ma, and
  Haibing Guan.
\newblock Fedala: Adaptive local aggregation for personalized federated
  learning.
\newblock In \emph{Proceedings of the AAAI Conference on Artificial
  Intelligence}, volume~37, pages 11237--11244, 2023.

\bibitem[Zhang et~al.(2017)Zhang, Guo, Dong, He, Xu, and Chen]{Zhang2017}
Shuyi Zhang, Bin Guo, Anlan Dong, Jing He, Ziping Xu, and Song~Xi Chen.
\newblock Cautionary tales on air-quality improvement in beijing.
\newblock \emph{Proceedings of the Royal Society A: Mathematical, Physical and
  Engineering Sciences}, 473\penalty0 (2205):\penalty0 20170457, 2017.

\end{thebibliography}

\appendix
\newpage
\section{Proof of Theorem \ref{th:1}} \label{A}
This section provides the proof of Theorem \ref{th:1}. The proof follows similar steps to those in \cite{Mitra2021}, and it is included here for the sake of completeness and to make the paper self-contained. 

According to \eqref{eq:3} and \eqref{eq:4}, for any $\vtheta$ it can be written that
\begin{align}
    \|\vtheta_{t+1} - \vtheta\|^2 =& \left\|\frac{1}{N}\sum_{i=1}^{N}{\boldsymbol{\phi}_{i,t+1}} - \vtheta \right\|^2 = \left\|\vtheta_t - \frac{\eta}{N} \sum_{i=1}^{N}{\nabla \gL(f(\vx_{i,t};\vtheta_t),y_{i,t}) } - \vtheta \right\|^2 \nonumber \\ =& \|\vtheta_t - \vtheta\|^2 + \left\|\frac{\eta}{N} \sum_{i=1}^{N}{\nabla \gL(f(\vx_{i,t};\vtheta_t),y_{i,t})}\right\|^2 \nonumber \\ &- \frac{2\eta}{N} \sum_{i=1}^{N}{(\vtheta_t - \vtheta)^\top \nabla \gL(f(\vx_{i,t};\vtheta_t),y_{i,t})}. \label{eq:1ap}
\end{align}
Due to convexity of $\gL(\cdot,\cdot)$ it can be concluded that 
\begin{align}
    \frac{2\eta}{N} \sum_{i=1}^{N}{\gL(f(\vx_{i,t};\vtheta_t),y_{i,t}) - \gL(f(\vx_{i,t};\vtheta),y_{i,t})} \le \frac{2\eta}{N} \sum_{i=1}^{N}{(\vtheta_t - \vtheta)^\top \nabla \gL(f(\vx_{i,t};\vtheta_t),y_{i,t})}. \label{eq:2ap}
\end{align}
Combining \eqref{eq:1ap} with \eqref{eq:2ap}, we get
\begin{align}
    & \frac{2\eta}{N} \sum_{i=1}^{N}{\gL(f(\vx_{i,t};\vtheta_t),y_{i,t})} - \frac{2\eta}{N}\sum_{i=1}^{N}{\gL(f(\vx_{i,t};\vtheta),y_{i,t})} \nonumber \\ \le &\|\vtheta_t - \vtheta\|^2 - \|\vtheta_{t+1} - \vtheta\|^2 + \left\| \frac{\eta}{N}\sum_{i=1}^{N}{\nabla \gL(f(\vx_{i,t};\vtheta_t),y_{i,t})} \right\|^2. \label{eq:3ap}
\end{align}
Using assumption A2 and Arithmetic Mean Geometric Mean (AM-GM) inequality it can be written that
\begin{align}
    \left\| \sum_{i=1}^{N}{\nabla \gL(f(\vx_{i,t};\vtheta_t),y_{i,t})}\right\|^2 \le N \sum_{i=1}^{N}{\left\| \nabla \gL(f(\vx_{i,t};\vtheta_t),y_{i,t})\right\|^2} \le N^2G^2. \label{eq:4ap}
\end{align}
Combining \eqref{eq:3ap} with \eqref{eq:4ap}, we arrive at
\begin{align}
    \frac{2\eta}{N} \sum_{i=1}^{N}{\gL(f(\vx_{i,t};\vtheta_t),y_{i,t})} - \frac{2\eta}{N}\sum_{i=1}^{N}{\gL(f(\vx_{i,t};\vtheta),y_{i,t})} \le \|\vtheta_t - \vtheta\|^2 - \|\vtheta_{t+1} - \vtheta\|^2 + \eta^2 G^2. \label{eq:5ap}
\end{align}
%Let $\rho$ be the probability that a client is chosen by the server to be in $\sC_t$. Since at each time step $t$, the server randomly chooses a subset of clients to use their calculated gradients to update the parameter $\vtheta_t$, parameter $\vtheta_t$ depends on randomization in client selection from time step $1$ to $t-1$. Moreover, since the server chooses a subset of clients uniformly at random every time step, $\1_{i \in \sC_t}$ can be viewed as a random variable whose value is independent of randomization in prior time steps. Therefore, it can be concluded that $\1_{i \in \sC_t}$ is independent of $\vtheta_t$. Taking the expectation from both sides of \eqref{eq:5ap} with respect to randomization in choosing clients from time step $1$ up until time step $t$, we obtain 
%\begin{align}
    %& \frac{2\eta \rho}{N} \sum_{i=1}^{N}{\E[\gL(f(\vx_{i,t};\vtheta_t),y_{i,t})]} - \frac{2\eta \rho}{N}\sum_{i=1}^{N}{\gL(f(\vx_{i,t};\vtheta),y_{i,t})} \nonumber \\ \le & \E[\|\vtheta_t - \vtheta\|^2 - \|\vtheta_{t+1} - \vtheta\|^2] + \rho \eta^2 G^2. \label{eq:58ap}
%\end{align}
Dividing both sides by $\frac{2\eta}{N}$ yields
\begin{align}
    \sum_{i=1}^{N}{\gL(f(\vx_{i,t};\vtheta_t),y_{i,t})} - \sum_{i=1}^{N}{\gL(f(\vx_{i,t};\vtheta),y_{i,t})} \le \frac{N (\|\vtheta_t - \vtheta\|^2 - \|\vtheta_{t+1} - \vtheta\|^2)}{2\eta} + \frac{\eta N}{2} G^2. \label{eq:59ap}
\end{align}
Summing \eqref{eq:59ap} over time, we obtain
\begin{align}
    & \sum_{t=1}^{T}{\sum_{i=1}^{N}{\gL(f(\vx_{i,t};\vtheta_t),y_{i,t})}} - \sum_{t=1}^{T}{\sum_{i=1}^{N}{\gL(f(\vx_{i,t};\vtheta),y_{i,t})}} \nonumber \\ \le & \frac{N(\|\vtheta_0 - \vtheta\|^2 - \|\vtheta_{T+1} - \vtheta\|^2)}{2\eta} + \frac{\eta N}{2}G^2T. \label{eq:6ap}
\end{align}
Plugging in $\vtheta = \vtheta^*$ and $\vtheta_0 = \boldsymbol{0}$ into \eqref{eq:6ap} and considering the fact that $\|\vtheta_{T+1} - \vtheta\|^2 \ge 0$, we find
\begin{align}
    \sum_{t=1}^{T}{\sum_{i=1}^{N}{\gL(f(\vx_{i,t};\vtheta_t),y_{i,t})}} - \sum_{t=1}^{T}{\sum_{i=1}^{N}{\gL(f(\vx_{i,t};\vtheta^*),y_{i,t})}} \le \frac{N\|\vtheta^*\|^2 }{2\eta} + \frac{\eta N}{2}G^2T \label{eq:7ap}
\end{align}
which proves the Theorem.

\section{Proof of Theorem \ref{th:2}} \label{B}
According to \eqref{eq:8}, we can write
\begin{align}
    \frac{\alpha_{i,t+1}+\beta_{i,t+1}}{\alpha_{i,t}+\beta_{i,t}} =& \frac{\alpha_{i,t}}{\alpha_{i,t}+\beta_{i,t}}\exp\left(-\eta_c \gL(f(\vx_{i,t};\vtheta_t),y_{i,t})\right) \nonumber \\ &+ \frac{\beta_{i,t}}{\alpha_{i,t}+\beta_{i,t}}\exp\left(-\eta_c \gL(f(\vx_{i,t};\boldsymbol{\phi}_{i,t}),y_{i,t})\right). \label{eq:8ap}
\end{align}
Using the inequality $e^{-x} \le 1-x+\frac{1}{2}x^2, \forall x \ge 0$, from \eqref{eq:8ap} we arrive at
\begin{align}
    \frac{\alpha_{i,t+1}+\beta_{i,t+1}}{\alpha_{i,t}+\beta_{i,t}} \le& \frac{\alpha_{i,t}}{\alpha_{i,t}+\beta_{i,t}}\left(1-\eta_c \gL(f(\vx_{i,t};\vtheta_t),y_{i,t})+\frac{\eta_c^2}{2}\gL^2(f(\vx_{i,t};\vtheta_t),y_{i,t})\right) \nonumber \\ &+ \frac{\beta_{i,t}}{\alpha_{i,t}+\beta_{i,t}}\left(1-\eta_c \gL(f(\vx_{i,t};\boldsymbol{\phi}_{i,t}),y_{i,t})+\frac{\eta_c^2}{2}\gL^2(f(\vx_{i,t};\boldsymbol{\phi}_{i,t}),y_{i,t})\right). \label{eq:9ap}
\end{align}
Taking the logarithm from both sides of \eqref{eq:9ap} and using the inequality $1+x\le e^x$, we get
\begin{align}
    \ln(\frac{\alpha_{i,t+1}+\beta_{i,t+1}}{\alpha_{i,t}+\beta_{i,t}}) \le& \frac{\alpha_{i,t}}{\alpha_{i,t}+\beta_{i,t}}\left(\!-\eta_c \gL(f(\vx_{i,t};\vtheta_t),y_{i,t})+\frac{\eta_c^2}{2}\gL^2(f(\vx_{i,t};\vtheta_t),y_{i,t})\!\right) \nonumber \\ &+ \frac{\beta_{i,t}}{\alpha_{i,t}\!+\!\beta_{i,t}}\left(\!-\eta_c \gL(f(\vx_{i,t};\boldsymbol{\phi}_{i,t}),y_{i,t})+\frac{\eta_c^2}{2}\gL^2(f(\vx_{i,t};\boldsymbol{\phi}_{i,t}),y_{i,t})\!\right). \label{eq:10ap}
\end{align}
Considering the assumption that $0\le \gL(f(\vx;\vtheta),y)\le 1$, $\forall \vx,\vtheta$, \eqref{eq:10ap} can be relaxed to
\begin{align}
    \ln(\frac{\alpha_{i,t+1}+\beta_{i,t+1}}{\alpha_{i,t}+\beta_{i,t}}) \le& \frac{\alpha_{i,t}}{\alpha_{i,t}+\beta_{i,t}}\left(-\eta_c \gL(f(\vx_{i,t};\vtheta_t),y_{i,t})\right) \nonumber \\ &+ \frac{\beta_{i,t}}{\alpha_{i,t}+\beta_{i,t}}\left(-\eta_c \gL(f(\vx_{i,t};\boldsymbol{\phi}_{i,t}),y_{i,t})\right) + \frac{\eta_c^2}{2}. \label{eq:11ap}
\end{align}
Summing \eqref{eq:11ap} over time, we obtain
\begin{align}
    \ln(\frac{\alpha_{i,T+1}+\beta_{i,T+1}}{\alpha_{i,1}+\beta_{i,1}}) \le& \frac{\alpha_{i,t}}{\alpha_{i,t}+\beta_{i,t}}\left(-\eta_c\sum_{t=1}^{T}{ \gL(f(\vx_{i,t};\vtheta_t),y_{i,t})}\right) \nonumber \\ &+ \frac{\beta_{i,t}}{\alpha_{i,t}+\beta_{i,t}}\left(-\eta_c \sum_{t=1}^{T}{\gL(f(\vx_{i,t};\boldsymbol{\phi}_{i,t}),y_{i,t})}\right) + \frac{\eta_c^2 T}{2}. \label{eq:12ap}
\end{align}
According to H\"{o}lder's inequality, for any positive real numbers $p$ and $q$ satisfying $\frac{1}{p} + \frac{1}{q}=1$, the following inequality holds:
\begin{align}
    \frac{\alpha_{i,T+1}}{p} + \frac{\beta_{i,T+1}}{q} \ge \alpha_{i,T+1}^\frac{1}{p} \beta_{i,T+1}^\frac{1}{q}. \label{eq:13ap}
\end{align}
To meet the condition $\frac{1}{p} + \frac{1}{q}=1$, it is necessary that $p\ge 1$ and $q \ge 1$. Consequently, \eqref{eq:13ap} can be modified to:
\begin{align}
    \alpha_{i,T+1} + \beta_{i,T+1} \ge \alpha_{i,T+1}^\frac{1}{p} \beta_{i,T+1}^\frac{1}{q}. \label{eq:14ap}
\end{align}
Considering the fact that $\alpha_{i,1}=\beta_{i,1}=1$, based on \eqref{eq:14ap} we can write
\begin{align}
    \ln(\frac{\alpha_{i,T+1}+\beta_{i,T+1}}{\alpha_{i,1}+\beta_{i,1}}) \ge \frac{1}{p} \ln(\alpha_{i,T+1}) + \frac{1}{q} \ln(\beta_{i,T+1}) - \ln(2). \label{eq:15ap}
\end{align}
According to the update rule in \eqref{eq:8ap}, \eqref{eq:15ap} is equivalent to
\begin{align}
    \ln(\frac{\alpha_{i,T+1}+\beta_{i,T+1}}{\alpha_{i,1}+\beta_{i,1}}) \ge -\frac{\eta_c}{p} \sum_{t=1}^{T}{\gL(f(\vx_{i,t};\vtheta_t),y_{i,t})} - \frac{\eta_c}{q} \sum_{t=1}^{T}{\gL(f(\vx_{i,t};\boldsymbol{\phi}_{i,t}),y_{i,t})} - \ln(2). \label{eq:16ap}
\end{align}
Combining \eqref{eq:12ap} with \eqref{eq:16ap} we arrive at
\begin{align}
    & \frac{\alpha_{i,t}}{\alpha_{i,t}+\beta_{i,t}}\sum_{t=1}^{T}{ \gL(f(\vx_{i,t};\vtheta_t),y_{i,t})} + \frac{\beta_{i,t}}{\alpha_{i,t}+\beta_{i,t}} \sum_{t=1}^{T}{\gL(f(\vx_{i,t};\boldsymbol{\phi}_{i,t}),y_{i,t})} \nonumber \\ &-\frac{1}{p} \sum_{t=1}^{T}{\gL(f(\vx_{i,t};\vtheta_t),y_{i,t})} - \frac{1}{q} \sum_{t=1}^{T}{\gL(f(\vx_{i,t};\boldsymbol{\phi}_{i,t}),y_{i,t})} \le \frac{\ln(2)}{\eta_c} + \frac{\eta_c T}{2}. \label{eq:17ap}
\end{align}
Due to the convexity of $\gL(\cdot,\cdot)$, we can write
\begin{align}
    \gL(f_{i,t}(\vx_{i,t}),y_{i,t}) &= \gL\left(\frac{\alpha_{i,t}}{\alpha_{i,t}+\beta_{i,t}}f(\vx_{i,t};\vtheta_t) + \frac{\beta_{i,t}}{\alpha_{i,t}+\beta_{i,t}}f(\vx_{i,t};\boldsymbol{\phi}_{i,t}),y_{i,t}\right) \nonumber \\ &\le \frac{\alpha_{i,t}}{\alpha_{i,t}+\beta_{i,t}}\sum_{t=1}^{T}{ \gL(f(\vx_{i,t};\vtheta_t),y_{i,t})} + \frac{\beta_{i,t}}{\alpha_{i,t}+\beta_{i,t}} \sum_{t=1}^{T}{\gL(f(\vx_{i,t};\boldsymbol{\phi}_{i,t}),y_{i,t})}. \label{eq:18ap} 
\end{align}
Combining \eqref{eq:17ap} with \eqref{eq:18ap} we get
\begin{align}
    & \sum_{t=1}^{T}{\gL(f_{i,t}(\vx_{i,t}),y_{i,t})} -\frac{1}{p} \sum_{t=1}^{T}{\gL(f(\vx_{i,t};\vtheta_t),y_{i,t})} - \frac{1}{q} \sum_{t=1}^{T}{\gL(f(\vx_{i,t};\boldsymbol{\phi}_{i,t}),y_{i,t})} \nonumber \\ \le& \frac{\ln(2)}{\eta_c} + \frac{\eta_c T}{2} \label{eq:19ap}
\end{align}
Substituting $p=\infty$ and $q=1$ in \eqref{eq:19ap}, we obtain
\begin{align}
    \sum_{t=1}^{T}{\gL(f_{i,t}(\vx_{i,t}),y_{i,t})} - \sum_{t=1}^{T}{\gL(f(\vx_{i,t};\boldsymbol{\phi}_{i,t}),y_{i,t})} \le \frac{\ln(2)}{\eta_c} + \frac{\eta_c T}{2}. \label{eq:20ap}
\end{align}
Since Fed-POE updates $\boldsymbol{\phi}_{i,t}$ locally using online gradient descent as outlined in \eqref{eq:6}, according to \eqref{eq:71ap}, for any $\boldsymbol{\phi}$ we can write
\begin{align}
    \sum_{t=1}^{T}{\gL(f(\vx_{i,t};\boldsymbol{\phi}_{i,t}),y_{i,t})} - \sum_{t=1}^{T}{\gL(f(\vx_{i,t};\boldsymbol{\phi}),y_{i,t})} \le \frac{\|\boldsymbol{\phi}\|^2 }{2\eta} + \frac{\eta}{2}G^2T. \label{eq:21ap}
\end{align}
Substituting $\boldsymbol{\phi}_i^*$ in \eqref{eq:21ap} along with combining \eqref{eq:20ap} with \eqref{eq:21ap}, we can conclude that
\begin{align}
    \sum_{t=1}^{T}{\gL(f_{i,t}(\vx_{i,t}),y_{i,t})} - \sum_{t=1}^{T}{\gL(f(\vx_{i,t};\boldsymbol{\phi}_i^*),y_{i,t})} \le \frac{\|\boldsymbol{\phi}_i^*\|^2 }{2\eta} + \frac{\ln(2)}{\eta_c} + \frac{\eta}{2}G^2T + \frac{\eta_c T}{2} \label{eq:22ap}
\end{align}
which proves the personalized regret upper bound of Fed-POE in \eqref{eq:9}. Furthermore, plunging in $p=1$ and $q=\infty$ into \eqref{eq:19ap}, we get
\begin{align}
    \sum_{t=1}^{T}{\gL(f_{i,t}(\vx_{i,t}),y_{i,t})} - \sum_{t=1}^{T}{\gL(f(\vx_{i,t};\vtheta_t),y_{i,t})} \le \frac{\ln(2)}{\eta_c} + \frac{\eta_c T}{2}. \label{eq:23ap}
\end{align}
Summing \eqref{eq:23ap} over all clients, we obtain
\begin{align}
    \sum_{t=1}^{T}{\sum_{i=1}^{N}{\gL(f_{i,t}(\vx_{i,t}),y_{i,t})}} - \sum_{t=1}^{T}{\sum_{i=1}^{N}{\gL(f(\vx_{i,t};\vtheta_t),y_{i,t})}} \le \frac{N\ln(2)}{\eta_c} + \frac{\eta_c NT}{2}. \label{eq:24ap}
\end{align}
Combining \eqref{eq:24ap} with \eqref{eq:7ap}, we arrive at
\begin{align}
    & \sum_{t=1}^{T}{\sum_{i=1}^{N}{\gL(f_{i,t}(\vx_{i,t}),y_{i,t})}} - \sum_{t=1}^{T}{\sum_{i=1}^{N}{\gL(f(\vx_{i,t};\vtheta^*),y_{i,t})}} \nonumber \\ \le & \frac{N\|\vtheta^*\|^2 }{2\eta} + \frac{N\ln(2)}{\eta_c} + \frac{\eta N}{2}G^2T + \frac{\eta_c NT}{2} \label{eq:25ap}
\end{align}
which proves the Theorem.

\section{Proof of Theorem \ref{th:3}} \label{C}
According to assumption A3 that $0 \le \gL(f(\vx;\vtheta),y) \le 1$, it can be written that
\begin{align}
    \frac{1}{N}\sum_{t=1}^{U}{\sum_{i=1}^{N}{\gL(f_{i,t}(\vx_{i,t}),y_{i,t})}} - \frac{1}{N}\sum_{t=1}^{U}{\sum_{i=1}^{N}{\gL(h^*(\vx_{i,t}),y_{i,t})}} \le U. \label{eq:26ap}
\end{align}
Let $\ell_{ij,t}$ denote the importance sampling loss estimate, which is expressed as
\begin{align}
    \ell_{ij,t} = \frac{\gL(f(\vx_{i,t};\boldsymbol{\rho}_j),y_{i,t})}{q_{ij,t}}\1_{j\in\sM_{i,t}}. \label{eq:27ap}
\end{align}
Let the total number of model parameters stored by the server after time step $U$ is $D$ and $W_{i,t}=\sum_{j=1}^{D}{w_{ij,t}}$. For any $t>U$, considering \eqref{eq:15}, we can write
\begin{align}
    \frac{W_{i,t+1}}{W_{i,t}} = \sum_{j=1}^{D}{\frac{w_{ij,t}}{W_{i,t}}\exp(-\eta_c \ell_{ij,t})}=\sum_{j=1}^{D}{p_{ij,t}\exp(-\eta_c \ell_{ij,t})}. \label{eq:28ap}
\end{align}
Employing the inequality $e^{-x} \le 1-x+\frac{1}{2}x^2, \forall x \ge 0$, from \eqref{eq:28ap} we obtain
\begin{align}
    \frac{W_{i,t+1}}{W_{i,t}} \le \sum_{j=1}^{D}{p_{ij,t}(1-\eta_c \ell_{ij,t}+\frac{\eta_c^2}{2} \ell_{ij,t}^2)}. \label{eq:29ap}
\end{align}
Taking the logarithm from both sides of \eqref{eq:29ap} and using the inequality $1+x\le e^x$, we arrive at
\begin{align}
    \ln \frac{W_{i,t+1}}{W_{i,t}} \le \sum_{j=1}^{D}{p_{ij,t}(-\eta_c \ell_{ij,t}+\frac{\eta_c^2}{2} \ell_{ij,t}^2)}. \label{eq:30ap}
\end{align}
Summing \eqref{eq:30ap} over time, we get
\begin{align}
    \ln \frac{W_{i,T+1}}{W_{i,U}} \le \sum_{t=U}^{T}{\sum_{j=1}^{D}{p_{ij,t}(-\eta_c \ell_{ij,t}+\frac{\eta_c^2}{2} \ell_{ij,t}^2)}}. \label{eq:31ap}
\end{align}
Considering the fact that the weights $\{w_{ij,t}\}_{j=1}^D$ are initialized as $w_{ij,1} = 1$, $\forall j \in [D]$, it can concluded that $W_{i,U} \le D$. Theefore, for any $j \in [D]$, the left hand side of \eqref{eq:31ap} is bounded from below as
\begin{align}
    \ln \frac{W_{i,T+1}}{W_{i,U}} \ge \ln \frac{w_{ij,T+1}}{W_{i,U}} \ge \ln \frac{w_{ij,T+1}}{D} = - \sum_{t=U}^{T}{\eta_c \ell_{ij,t}} - \ln D. \label{eq:32ap}
\end{align}
Combining \eqref{eq:32ap} with \eqref{eq:31ap} yields
\begin{align}
    \sum_{t=U}^{T}{\sum_{j=1}^{D}{p_{ij,t}\ell_{ij,t}}} - \sum_{t=U}^{T}{\ell_{ij,t}} \le \frac{\ln D}{\eta_c} + \frac{\eta_c}{2} \sum_{t=U}^{T}{\sum_{j=1}^{D}{p_{ij,t}\ell_{ij,t}^2}}. \label{eq:33ap}
\end{align}
The expected values of $\ell_{ij,t}$ and $\ell_{ij,t}^2$ given observed losses up until time step $t$ can be obtained as
\begin{subequations} \label{eq:34ap}
    \begin{align}
        \E_t[\ell_{ij,t}] &= \frac{\gL(f(\vx_{i,t};\boldsymbol{\rho}_j),y_{i,t})}{q_{ij,t}}\E_t[\1_{j\in\sM_{i,t}}] = \gL(f(\vx_{i,t};\boldsymbol{\rho}_j),y_{i,t}) \label{eq:34apa} \\
        \E_t[\ell_{ij,t}^2] &= \frac{\gL(f(\vx_{i,t};\boldsymbol{\rho}_j),y_{i,t})^2}{q_{ij,t}^2}\E_t[\1_{j\in\sM_{i,t}}] = \frac{\gL(f(\vx_{i,t};\boldsymbol{\rho}_j),y_{i,t})^2}{q_{ij,t}} \le \frac{1}{q_{ij,t}}. \label{eq:34apb}
    \end{align}
\end{subequations}
Taking the expectation from \eqref{eq:33ap}, we arrive at
\begin{align}
    \sum_{t=U}^{T}{\sum_{j=1}^{D}{p_{ij,t}\gL(f(\vx_{i,t};\boldsymbol{\rho}_j),y_{i,t})}} - \sum_{t=U}^{T}{\gL(f(\vx_{i,t};\boldsymbol{\rho}_j),y_{i,t})} \le \frac{\ln D}{\eta_c} + \frac{\eta_c}{2} \sum_{t=U}^{T}{\sum_{j=1}^{D}{\frac{p_{ij,t}}{q_{ij,t}}}}. \label{eq:35ap}
\end{align}
Since $q_{ij,t} = 1-(1-p_{ij,t})^M = p_{ij,t}(1+(1-p_{ij,t})+(1-p_{ij,t})^2+\ldots+(1-p_{ij,t})^{M-1})$, it can be concluded that $q_{ij,t}\ge p_{ij,t}$. Therefore, \eqref{eq:35ap} can be relaxed to
\begin{align}
    \sum_{t=U}^{T}{\sum_{j=1}^{D}{p_{ij,t}\gL(f(\vx_{i,t};\boldsymbol{\rho}_j),y_{i,t})}} - \sum_{t=U}^{T}{\gL(f(\vx_{i,t};\boldsymbol{\rho}_j),y_{i,t})} \le \frac{\ln D}{\eta_c} + \frac{\eta_c}{2} D(T-U). \label{eq:36ap}
\end{align}
According to model selection procedure adopted by Fed-POE presented in \Algref{alg:2}, client $i$ chooses a subset of models by sampling them in $M$ rounds with replacement. Let $a_{ij,t} \ge 0$ denote the number of times that the model $j$ in $\sD_t$ is chosen by client $i$ at time step $t$. The number of different situations for selected subset of models $\sM_{i,t}$ is equal to the number of solutions for the linear equation
\begin{align}
    a_{i1,t} + \ldots + a_{iD,t} = M, a_{ij,t}\ge 0, \forall j\in [D]. \label{eq:37ap}
\end{align}
Let $\sA$ denote the set of all possible solutions for \eqref{eq:37ap} such that if $\va \in \sA$ where $\va=[a_1,\ldots,a_D]$, $a_1,\ldots,a_D$ satisfies \eqref{eq:37ap}. Therefore, for the expected loss of the ensemble $\Tilde{f}_{i,t}(\vx_{i,t})$ in \eqref{eq:20}, we can write
\begin{align}
    \E_t[\gL(\Tilde{f}_{i,t}(\vx_{i,t}),y_{i,t})] = \sum_{k=1}^{|\sA|}{\prod_{j=1}^{D}{p_{ij,t}^{a_{j,k}}}\gL(\Tilde{f}_{i,t}^{(k)}(\vx_{i,t}),y_{i,t})} \label{eq:38ap}
\end{align}
where $\Tilde{f}_{i,t}^{(k)}(\vx_{i,t})$ denote the $k$-th possible ensemble model generated by client $i$ using Fed-POE. Using the Jensen inequality and convexity of the loss function, we can relax \eqref{eq:38ap} to
\begin{align}
    \E_t[\gL(\Tilde{f}_{i,t}(\vx_{i,t}),y_{i,t})] &= \sum_{k=1}^{|\sA|}{\left(\prod_{j=1}^{D}{p_{ij,t}^{a_{j,k}}}\right)\gL(\Tilde{f}_{i,t}^{(k)}(\vx_{i,t}),y_{i,t})} \nonumber \\ &\le \sum_{k=1}^{|\sA|}{\left(\prod_{j=1}^{D}{p_{ij,t}^{a_{j,k}}}\right)\sum_{m \in \sM_{i,t}^{(k)}}{\frac{w_{im,t}}{W_{i,t}^{(k)}}\gL(f(\vx_{i,t};\boldsymbol{\rho}_j),y_{i,t})}} \label{eq:39ap}
\end{align}
where $\sM_{i,t}^{(k)}$ and $W_{i,t}^{(k)}$ are the $k$-th possible model subset and weight summations, respectively. Rearranging the right hand side of \eqref{eq:39ap}, we can write
\begin{align}
    \E_t[\gL(\Tilde{f}_{i,t}(\vx_{i,t}),y_{i,t})] \le \sum_{j=1}^{D}{p_{ij,t}\sum_{k=1}^{|\sB_j|}{\left(\prod_{m=1}^{D}{p_{im,t}^{b_{m,k}}}\right)\frac{w_{ij,t}}{W_{i,t}^{(k)}}\gL(f(\vx_{i,t};\boldsymbol{\rho}_j),y_{i,t})}} \label{eq:40ap}
\end{align}
where $\sB_j$ is the set of all possible solutions for the linear equation in \eqref{eq:37ap} condition on $a_{ij,t} \ge 1$. Since for any $k$, we have $w_{ij,t} \le W_{i,t}^{(k)}$, \eqref{eq:40ap} can be relaxed to
\begin{align}
    \E_t[\gL(\Tilde{f}_{i,t}(\vx_{i,t}),y_{i,t})] \le \sum_{j=1}^{D}{p_{ij,t}\sum_{k=1}^{|\sB_j|}{\left(\prod_{m=1}^{D}{p_{im,t}^{b_{m,k}}}\right)\gL(f(\vx_{i,t};\boldsymbol{\rho}_j),y_{i,t})}}. \label{eq:41ap}
\end{align}
Since $\sum_{k=1}^{|\sB_j|}{\left(\prod_{m=1}^{D}{p_{im,t}^{b_{m,k}}}\right)}$ includes all possibilities in $\sB_j$, we can conclude that $\sum_{k=1}^{|\sB_j|}{\left(\prod_{m=1}^{D}{p_{im,t}^{b_{m,k}}}\right)}=1$. Combining this with \eqref{eq:41ap}, we obtain
\begin{align}
    \E_t[\gL(\Tilde{f}_{i,t}(\vx_{i,t}),y_{i,t})] \le \sum_{j=1}^{D}{p_{ij,t}\gL(f(\vx_{i,t};\boldsymbol{\rho}_j),y_{i,t})}. \label{eq:42ap}
\end{align}
Combining \eqref{eq:42ap} with \eqref{eq:36ap}, we arrive at
\begin{align}
    \sum_{t=U}^{T}{\E_t[\gL(\Tilde{f}_{i,t}(\vx_{i,t}),y_{i,t})]} - \sum_{t=U}^{T}{\gL(f(\vx_{i,t};\boldsymbol{\rho}_j),y_{i,t})} \le \frac{\ln D}{\eta_c} + \frac{\eta_c}{2} D(T-U). \label{eq:43ap}
\end{align}
Since $0 \le \gL(f(\vx;\vtheta),y) \le 1$, it can be written that
\begin{align}
    \sum_{t=1}^{U}{\E_t[\gL(\Tilde{f}_{i,t}(\vx_{i,t}),y_{i,t})]} - \sum_{t=1}^{U}{\gL(f(\vx_{i,t};\boldsymbol{\rho}_j),y_{i,t})} \le U. \label{eq:44ap}
\end{align}
Combining \eqref{eq:44ap} with \eqref{eq:43ap}, we get
\begin{align}
    \sum_{t=1}^{T}{\E_t[\gL(\Tilde{f}_{i,t}(\vx_{i,t}),y_{i,t})]} - \sum_{t=1}^{T}{\gL(f(\vx_{i,t};\boldsymbol{\rho}_j),y_{i,t})} \le \frac{\ln D}{\eta_c} + \frac{\eta_c}{2} D(T-U) + U. \label{eq:45ap}
\end{align}
Since $\bar{f}_{i,t}(\cdot)$ similar to $f_{i,t}(\cdot)$ is the ensemble of two models, following the same derivation steps from \eqref{eq:8ap} to \eqref{eq:20ap} by substituting $\bar{f}_{i,t}(\vx_{i,t})$, $f_{i,t}(\vx_{i,t})$ and $\Tilde{f}_{i,t}(\vx_{i,t})$ with $f_{i,t}(\vx_{i,t})$, $f(\vx_{i,t};\vtheta_t)$ and $f(\vx_{i,t};\boldsymbol{\phi}_{i,t})$, respectively, we can conclude that
\begin{subequations} \label{eq:46ap}
    \begin{align}
        \sum_{t=1}^{T}{\gL(\bar{f}_{i,t}(\vx_{i,t}),y_{i,t})} - \sum_{t=1}^{T}{\gL(f_{i,t}(\vx_{i,t}),y_{i,t})} \le& \frac{\ln(2)}{\eta_c} + \frac{\eta_c T}{2}, \label{eq:46apa} \\
        \sum_{t=1}^{T}{\gL(\bar{f}_{i,t}(\vx_{i,t}),y_{i,t})} - \sum_{t=1}^{T}{\gL(\Tilde{f}_{i,t}(\vx_{i,t}),y_{i,t})} \le& \frac{\ln(2)}{\eta_c} + \frac{\eta_c T}{2}. \label{eq:46apb}
    \end{align}
\end{subequations}
Taking the expectation from both sides of \eqref{eq:46apb} with respect to randomization in model selection along with combining \eqref{eq:46apb} with \eqref{eq:45ap} yields
\begin{align}
    &\sum_{t=1}^{T}{\E_t[\gL(\bar{f}_{i,t}(\vx_{i,t}),y_{i,t})]} - \sum_{t=1}^{T}{\gL(f(\vx_{i,t};\boldsymbol{\rho}_j),y_{i,t})} \nonumber \\ \le& \frac{\ln 2D}{\eta_c} + \frac{\eta_c}{2} (D+1)T + (1-\frac{\eta_c}{2}D)U. \label{eq:47ap}
\end{align}
Furthermore, combining \eqref{eq:46apa} with \eqref{eq:20ap} and \eqref{eq:23ap} and taking the expectation with respect to model selection randomization, we obtain
\begin{subequations} \label{eq:48ap}
    \begin{align}
        \sum_{t=1}^{T}{\E_t[\gL(\bar{f}_{i,t}(\vx_{i,t}),y_{i,t})]} - \sum_{t=1}^{T}{\gL(f(\vx_{i,t};\boldsymbol{\phi}_{i,t}),y_{i,t})} \le& \frac{\ln(4)}{\eta_c} + \eta_c T, \label{eq:48apa} \\
        \sum_{t=1}^{T}{\E_t[\gL(\bar{f}_{i,t}(\vx_{i,t}),y_{i,t})]} - \sum_{t=1}^{T}{\gL(f(\vx_{i,t};\vtheta_t),y_{i,t})} \le& \frac{\ln(4)}{\eta_c} + \eta_c T. \label{eq:48apb}
    \end{align}
\end{subequations}
Recall that $h_j(\cdot)$ associated with the $j$-th model parameter in $\sD_t$ be defined as $h_j(\vx_{i,t})=f(\vx_{i,t};\boldsymbol{\rho}_j)$ while $h_{\text{loc}}(\cdot)$ and $h_{\text{fed}}(\cdot)$ correspond to the local and federated models, respectively, defined as $h_{\text{loc}}(\vx_{i,t})=f(\vx_{i,t};\boldsymbol{\phi}_{i,t})$ and $h_{\text{fed}}(\vx_{i,t})=f(\vx_{i,t};\vtheta_t)$. Also recall that $\sH:=\{h_j \mid \forall j: 1 \le j \le |\sD_T|\} \cup \{h_{\text{loc}}, h_{\text{fed}}\}$. Comparing the right hand sides of \eqref{eq:48apa} and \eqref{eq:48apb} with that of \eqref{eq:47ap} and considering the fact that $D \ge 2$, for any $h \in \sH$ we can write
\begin{align}
    \sum_{t=1}^{T}{\E_t[\gL(\bar{f}_{i,t}(\vx_{i,t}),y_{i,t})]} - \sum_{t=1}^{T}{\gL(h(\vx_{i,t}),y_{i,t})} \le \frac{\ln 2D}{\eta_c} + \frac{\eta_c}{2} (D+1)T + (1-\frac{\eta_c}{2}D)U. \label{eq:49ap}
\end{align}
By substituting $h(\cdot)$ with $h_i^*(\cdot)$ as defined in \eqref{eq:16b} and considering the fact that $D \le U/n$, we obtain the personalized regret upper bound for client $i$ as shown in \eqref{eq:21}. Moreover, taking the average of \eqref{eq:49ap} across clients and substituting $h(\cdot)$ with $h^*(\cdot)$, we arrive at
\begin{align}
    &\frac{1}{N}\sum_{t=1}^{T}{\sum_{i=1}^{N}{\E_t[\gL(\bar{f}_{i,t}(\vx_{i,t}),y_{i,t})]}} - \frac{1}{N}\sum_{t=1}^{T}{\sum_{i=1}^{N}{\gL(h^*(\vx_{i,t}),y_{i,t})}} \nonumber \\ \le& \frac{\ln 2D}{\eta_c} + \frac{\eta_c}{2} (D+1)T + (1-\frac{\eta_c}{2}D)U \label{eq:50ap}
\end{align}
which proves the theorem.

\section{Supplementary Experimental Details} \label{D}
This section presents supplementary experimental results and details about experimental setup. All experiments were carried out using Intel(R) Core(TM) i7-10510U CPU @ 1.80 GHz 2.30 GHz processor with a 64-bit {Windows} operating system. 

\subsection{Regression Data Distribution}
As it is pointed out in section \ref{sec:exp}, the present paper tests the performance of algorithms for online regression task on Air and WEC datasets. These datasets are downloaded from UCI Machine Learning Repository \cite{Kelly2023}. Each data sample in Air dataset includes air quality information such as concentration of some chemicals in the air. Data samples in Air dataset collected from four different geographical locations. Moreover, data samples in WEC, collected from wave energy converters in four different geographical locations. In order to distribute data, clients are partitioned into $4$ groups. For each group, $70\%$ of data samples observed by each client in the group belongs to a specific geographical location while $10\%$ of observed data samples belong to each of the rest $3$ locations.

\subsection{Random Feature Kernel-based Models}
As it is pointed in section \ref{sec:exp}, the proposed Fed-POE and all baselines utilize a random feature kernel-based model to perform online regression task. In what follows we explain random feature-based kernel models. Let $\kappa(\cdot,\cdot)$ be a positive-definite function called kernel such that $\kappa(\vx,\vx^\prime)$ measures the similarity between $\vx$ and $\vx^\prime$. In online kernel learning context, at time step $t+1$, the following prediction is made for $\vx$ (see e.g. \citep{Wahba1990,Hoi2013,Sahoo2014}):
\begin{align}
    f_\kappa(\vx;\boldsymbol{\alpha}_t) = \sum_{\tau=1}^{t}{\sum_{i=1}^{N}{\alpha_{i,\tau}\kappa(\vx,\vx_{i,\tau})}} \label{eq:51ap}
\end{align}
where $\boldsymbol{\alpha}_t = [\alpha_{1,1},\ldots,\alpha_{N,1},\ldots,\alpha_{1,t},\ldots,\alpha_{N,t}]$ denotes the learnable parameters. Therefore, the number of parameters that should be learned grows with time. In order to alleviate the computational complexity of online kernel learning, random feature approximation \citep{Rahimi2007} can be employed. In fact, using random feature approximation, the number of parameters that needs be learned is time-invariant and is selected by the algorithm.
Assume that $\kappa(\cdot)$ is a shift-invariant kernel function such that $\kappa(\vx,\vx^\prime) = \kappa(\vx-\vx^\prime)$. Also, suppose that $\kappa(\cdot)$ is scaled such that $\kappa(\boldsymbol{0}) = 1$. Let $\xi(\cdot)$ denotes the Fourier transform of $\kappa(\cdot)$. According to definition of inverse Fourier transform $\kappa(\boldsymbol{0}) = \int_{-\infty}^{\infty}{\xi(\boldsymbol{\omega})d\boldsymbol{\omega}} = 1$. Therefore, it can be concluded that $\xi(\cdot)$ is a probability density function (PDF). Let $\boldsymbol{\omega}_1, \ldots, \boldsymbol{\omega}_D$ be drawn randomly from $\xi(\cdot)$ and called random features. Using the random features $\boldsymbol{\omega}_1, \ldots, \boldsymbol{\omega}_D$, the representation $\vz(\vx)$ is defined as
\begin{align}
    \vz(\vx) = \frac{1}{\sqrt{D}}[\sin(\boldsymbol{\omega}_1^\top \vx),\ldots,\sin(\boldsymbol{\omega}_D^\top \vx), \cos(\boldsymbol{\omega}_1^\top \vx),\ldots,\cos(\boldsymbol{\omega}_D^\top \vx)]. \label{eq:52ap}
\end{align}
Given random features $\boldsymbol{\omega}_1, \ldots, \boldsymbol{\omega}_D$ and using the proposed Fed-POE, at time step $t$, client $i$ makes prediction $f(\vx_{i,t};\vtheta_t) = \vtheta_t^\top \vz(\vx_{i,t})$. Clients and the server employ the proposed Fed-POE to learn the parameter $\vtheta_t$. As it can be inferred since the model $f(\cdot;\cdot)$ is linear with respect to model parameter $\vtheta_t$, using convex loss functions, the loss $\gL(f(\vx_{i,t};\vtheta_t),y_{i,t})$ is convex as well.

Furthermore, in section \ref{sec:exp}, the proposed Fed-POE utilizes three Gaussian kernels for online regression on Air and WEC datasets. In order to implement multi-kernel learning for Fed-POE, the prediction of kernels are linearly combined and the weights for linear combination is learned locally by each client. Let $f_{0.1}(\vx_{i,t})$, $f_{1}(\vx_{i,t})$ and $f_{10}(\vx_{i,t})$ represent predictions of Gaussian kernels with variances of $0.1$, $1$ and $10$, respectively. Then at time step $t$, client $i$ makes prediction $w_{0.1,it}f_{0.1}(\vx_{i,t}) + w_{1,it}f_{1}(\vx_{i,t}) + w_{10,it}f_{10}(\vx_{i,t})$. In order to update weights $w_{0.1,it}$, $w_{1,it}$ and $w_{10,it}$, client $i$ employs multiplicative update rule. As an example after observing the loss $\gL(f_{1}(\vx_{i,t}),y_{i,t})$, client $i$ updates $w_{1,it}$ as $w_{1,i(t+1)} = w_{1,it}\exp(-\gamma_i \gL(f_{1}(\vx_{i,t}),y_{i,t}))$ where $\gamma_i$ is a learning rate specified by client $i$.

\subsection{Image Classification Experimental Setup}
The pre-trained CNN used by Fed-POE and other baselines is biased toward class label $0$. For CIFAR-10, the pre-trained CNN is trained on a subset of the CIFAR-10 training data, consisting of $5,000$ samples with label $0$ and $500$ samples from each of the other $9$ class labels. For FMNIST, the model is trained on a subset of the FMNIST training data, consisting of $6,000$ samples with label $0$ and $500$ samples from each of the other $9$ class labels. The CNN models are trained using Tensorflow 2.16.1. We used the SGD optimizer with a learning rate of $10^{-3}$ and momentum of $0.9$. The models for CIFAR-10 and FMNIST were trained for $100$ epochs and $10$ epochs, respectively.

Clients receive test data samples sequentially and make prediction for the newly received sample. To distribute test data samples of CIFAR-10 among clients, we split clients into $10$ groups. For CIFAR-10, $55\%$ of samples observed by a client belongs to a specific class label while only $5\%$ of received samples belong to each of other $9$ class labels. For FMNIST, client data distribution is time-variant. Since the number of test samples is $10,000$ and the number of clients is $20$, it can be concluded that time horizon $T$ is $500$. In the first half of time steps (i.e. $t \le 250$), each client observes $200$ samples from the first $5$ class labels and $50$ samples from other $5$ class labels.  In the second half, this is reversed: clients observe $200$ samples from the last $5$ class labels and $50$ samples from the other class labels. In each half, each client is biased toward one of the five majority classes. For example, if a client is biased toward class $0$ in the first half, it observes $100$ samples from class label $0$, $25$ samples from each of class labels $1$ to $4$, and $10$ samples from each of class labels $5$ to $9$. Therefore, in each half, each client observes $100$ samples from one class, $25$ samples from each of four other classes, and $10$ samples from each of the remaining five classes.

To implement Fed-OMD for both regression and image classification, we used the $\ell_2$-norm as a regularizer function for mirror descent. For implementing Ditto for both regression and image classification, we set the regularization factor \(\lambda\) to 1. In the case of image classification using Fed-Rep, clients locally fine-tune the last two layers of the CNN model, while the rest of the network is used as the global backbone to generate representations. Furthermore, Fed-POE is compatible with any federated learning method and can utilize any federated algorithm. For CIFAR-10, Fed-POE uses Fed-OMD, while for FMNIST, Fed-POE uses Fed-Rep. To fine-tune the CNN model, Fed-POE and all baselines use the SGD optimizer and the cross-entropy loss function.

\subsection{Supplementary Results}
\begin{figure} [t]
\centering
\subfigure[CIFAR-10]{%
  \centering
  \includegraphics[scale=.4]{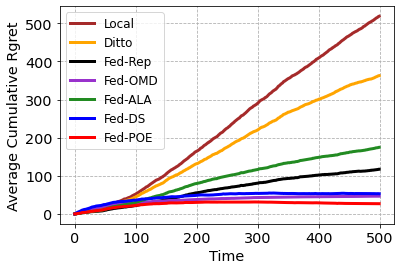}
    }
\quad
\subfigure[WEC]{%
\centering
  \includegraphics[scale=.4]{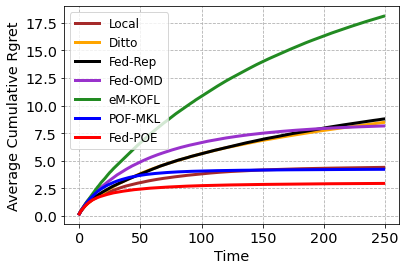}
    }
\caption{Cumulative regret over time on CIFAR-10 and WEC datasets.}
\label{fig:1}
\end{figure}

Table \ref{table:2} presents the accuracy of clients for image classification using Fed-POE with varying values of \(M\). As can be seen for both CIFAR-10 and FMNIST, when \(M>0\), the accuracy is higher than in the case where \(M=0\). The case where \(M=0\) corresponds to clients not using models stored by the server in their ensemble. Therefore, these results show that constructing the ensemble using previous federated model parameters stored by the server improves the accuracy of Fed-POE. This indicates the effectiveness of the model selection procedure of Fed-POE presented in \Algref{alg:2}. Additionally, these results show that increasing \(M\) does not necessarily lead to further accuracy improvement. This implies the effectiveness of Fed-POE's model selection in pruning model parameters from the ensemble that have relatively lower accuracy. Figure \ref{fig:1} illustrates the average cumulative global regret of clients over time using Fed-POE and all other baselines. As depicted, Fed-POE achieves sublinear regret for both CIFAR-10 and WEC datasets. This corroborates the theoretical results in Theorems \ref{th:2} and \ref{th:3}.

% \begin{figure} [t]
% \centering
% \subfigure[CIFAR-10]{%
%   \centering
%   \includegraphics[scale=.4]{regret_cifar10.png}
%     }
% \quad
% \subfigure[MNIST]{%
% \centering
%   \includegraphics[scale=.4]{regret_mnist.png}
%     }
% \caption{Cumulative regret on image classification with the change in the number of clients with fixed $NT = 10,000$.}
% \label{fig:1}
% \end{figure}

% \begin{figure} [t]
% \centering
% \subfigure[Air]{%
%   \centering
%   \includegraphics[scale=.4]{regret_air.png}
%     }
% \quad
% \subfigure[WEC]{%
% \centering
%   \includegraphics[scale=.4]{regret_wec.png}
%     }
% \caption{Cumulative regret on regression with the change in the number of clients with fixed $NT=100,000$.}
% \label{fig:2}
% \end{figure}

\begin{table}[t]
\setlength{\tabcolsep}{4pt}
\caption{Average accuracy and standard deviation across clients employing Fed-POE for image classification with varying values of $M$}
\label{table:2}
\begin{center}
\begin{tabular}{l||c|c|c|c}
\toprule
{\bf Datasets}    &$M=0$ &$M=4$ &$M=8$ &$M=16$
\\ \hline
CIFAR-10  &$65.55\% \pm 8.77\%$   &$66.50\% \pm 8.00\%$ &$\mathbf{66.54}\% \pm \mathbf{8.08}\%$   &$66.46\% \pm 7.98\%$ \\
FMNIST  &$79.03\% \pm 1.76\%$   &$79.12\% \pm 1.87\%$ &$\mathbf{79.23}\% \pm \mathbf{1.88}\%$   &$79.18\% \pm 1.85\%$ \\
\bottomrule
\end{tabular}
\end{center}
\end{table}

%%%%%%%%%%%%%%%%%%%%%%%%%%%%%%%%%%%%%%%%%%%%%%%%%%%%%%%%%%%%

\newpage
\section*{NeurIPS Paper Checklist}

\begin{enumerate}

\item {\bf Claims}
    \item[] Question: Do the main claims made in the abstract and introduction accurately reflect the paper's contributions and scope?
    \item[] Answer: \answerYes{} % Replace by \answerYes{}, \answerNo{}, or \answerNA{}.
    \item[] Justification: The abstract and introduction reflect that the contribution of this paper is the proposal of a novel personalized federated learning algorithm for online prediction and model fine-tuning.

\item {\bf Limitations}
    \item[] Question: Does the paper discuss the limitations of the work performed by the authors?
    \item[] Answer: \answerYes{} % Replace by \answerYes{}, \answerNo{}, or \answerNA{}.
    \item[] Justification: In Section \ref{OFRL}, the paper makes it clear that Fed-POE utilizes multiple models instead of a single model for prediction. This requires more computations than the case where clients use only one model. Computational efficiency of Fed-POE is discussed in Section \ref{OFRL}.

\item {\bf Theory Assumptions and Proofs}
    \item[] Question: For each theoretical result, does the paper provide the full set of assumptions and a complete (and correct) proof?
    \item[] Answer: \answerYes{} % Replace by \answerYes{}, \answerNo{}, or \answerNA{}.
    \item[] Justification: The paper provides the full set of assumptions in Section \ref{OFGD} and the complete proofs of all theorems can be found in Appendices \ref{A}, \ref{B} and \ref{C}.

    \item {\bf Experimental Result Reproducibility}
    \item[] Question: Does the paper fully disclose all the information needed to reproduce the main experimental results of the paper to the extent that it affects the main claims and/or conclusions of the paper (regardless of whether the code and data are provided or not)?
    \item[] Answer: \answerYes{} % Replace by \answerYes{}, \answerNo{}, or \answerNA{}.
    \item[] Justification: In Section \ref{sec:exp} and Appendix \ref{D}, we provide detailed information needed to reproduce the results.

\item {\bf Open access to data and code}
    \item[] Question: Does the paper provide open access to the data and code, with sufficient instructions to faithfully reproduce the main experimental results, as described in supplemental material?
    \item[] Answer: \answerYes{} % Replace by \answerYes{}, \answerNo{}, or \answerNA{}.
    \item[] Justification: We provide open access to the data and code.

\item {\bf Experimental Setting/Details}
    \item[] Question: Does the paper specify all the training and test details (e.g., data splits, hyperparameters, how they were chosen, type of optimizer, etc.) necessary to understand the results?
    \item[] Answer: \answerYes{} % Replace by \answerYes{}, \answerNo{}, or \answerNA{}.
    \item[] Justification: In Section \ref{sec:exp} and Appendix \ref{D}, we provide all the training and test details necessary to understand the results.

\item {\bf Experiment Statistical Significance}
    \item[] Question: Does the paper report error bars suitably and correctly defined or other appropriate information about the statistical significance of the experiments?
    \item[] Answer: \answerYes{} % Replace by \answerYes{}, \answerNo{}, or \answerNA{}.
    \item[] Justification: In Tables \ref{table:3}, \ref{table:1} and \ref{table:2}, We reported the standard deviation of the results.

\item {\bf Experiments Compute Resources}
    \item[] Question: For each experiment, does the paper provide sufficient information on the computer resources (type of compute workers, memory, time of execution) needed to reproduce the experiments?
    \item[] Answer: \answerYes{} % Replace by \answerYes{}, \answerNo{}, or \answerNA{}.
    \item[] Justification: In Appendix \ref{D}, we clarify that all experiments were carried out using Intel(R) Core(TM) i7-10510U CPU @ 1.80 GHz 2.30 GHz processor with a 64-bit {Windows} operating system.
    
\item {\bf Code Of Ethics}
    \item[] Question: Does the research conducted in the paper conform, in every respect, with the NeurIPS Code of Ethics \url{https://neurips.cc/public/EthicsGuidelines}?
    \item[] Answer: \answerYes{} % Replace by \answerYes{}, \answerNo{}, or \answerNA{}.
    \item[] Justification: We believe the research conducted in the paper conform, in every respect, with the NeurIPS Code of Ethics.

\item {\bf Broader Impacts}
    \item[] Question: Does the paper discuss both potential positive societal impacts and negative societal impacts of the work performed?
    \item[] Answer: \answerNA{} % Replace by \answerYes{}, \answerNo{}, or \answerNA{}.
    \item[] Justification: This paper studies the problem of personalized federated learning. While personalized federated learning, in general, has societal impact, we do not foresee any specific societal impact resulting from our work.
    
\item {\bf Safeguards}
    \item[] Question: Does the paper describe safeguards that have been put in place for responsible release of data or models that have a high risk for misuse (e.g., pretrained language models, image generators, or scraped datasets)?
    \item[] Answer: \answerNA{} % Replace by \answerYes{}, \answerNo{}, or \answerNA{}.
    \item[] Justification: We believe that the paper poses no such risks.

\item {\bf Licenses for existing assets}
    \item[] Question: Are the creators or original owners of assets (e.g., code, data, models), used in the paper, properly credited and are the license and terms of use explicitly mentioned and properly respected?
    \item[] Answer: \answerYes{} % Replace by \answerYes{}, \answerNo{}, or \answerNA{}.
    \item[] Justification: In Section \ref{sec:exp} and Appendix \ref{D}, we properly cited the original owners of assets.

\item {\bf New Assets}
    \item[] Question: Are new assets introduced in the paper well documented and is the documentation provided alongside the assets?
    \item[] Answer: \answerNA{} % Replace by \answerYes{}, \answerNo{}, or \answerNA{}.
    \item[] Justification: The paper does not release new assets.

\item {\bf Crowdsourcing and Research with Human Subjects}
    \item[] Question: For crowdsourcing experiments and research with human subjects, does the paper include the full text of instructions given to participants and screenshots, if applicable, as well as details about compensation (if any)? 
    \item[] Answer: \answerNA{} % Replace by \answerYes{}, \answerNo{}, or \answerNA{}.
    \item[] Justification: The paper does not involve crowdsourcing nor research with human subjects.

\item {\bf Institutional Review Board (IRB) Approvals or Equivalent for Research with Human Subjects}
    \item[] Question: Does the paper describe potential risks incurred by study participants, whether such risks were disclosed to the subjects, and whether Institutional Review Board (IRB) approvals (or an equivalent approval/review based on the requirements of your country or institution) were obtained?
    \item[] Answer: \answerNA{} % Replace by \answerYes{}, \answerNo{}, or \answerNA{}.
    \item[] Justification: The paper does not involve crowdsourcing nor research with human subjects.

\end{enumerate}

\end{document}